\documentclass[journal]{IEEEtran}

\ifCLASSINFOpdf
\else
\fi
\usepackage{times}
\usepackage{epsfig}
\usepackage{graphicx}
\usepackage{amsmath}
\usepackage{amssymb}
\usepackage{diagbox}
\usepackage{color}
\usepackage{makecell}
\usepackage{multirow}
\usepackage{subfigure}
\usepackage{url}
\usepackage[normalem]{ulem}
\renewcommand{\mathbf}{\boldsymbol}

\begin{document}
\title{Part-Aware Fine-grained Object Categorization using Weakly Supervised Part Detection Network}

\author{Yabin~Zhang,
        Kui~Jia,
        and~Zhixin~Wang
\thanks{This work was supported in part by the National Natural Science Foundation of China (Grant No.: 61771201), and the Program for Guangdong Introducing Innovative and Enterpreneurial Teams (Grant No.: 2017ZT07X183). 

Y. Zhang, K. Jia, and Z. Wang are with the School of Electronic and Information Engineering, South China University of Technology, Guangzhou, China. E-mails: zhang.yabin@mail.scut.edu.cn, kuijia@scut.edu.cn, wang.zhixin@mail.scut.edu.cn.
%\textit{* Correspondence to Kui~Jia.}
}

}


\markboth{IEEE TRANSACTIONS ON MULTIMEDIA,~Vol.~X, No.~X, August~2019}%
{Shell \MakeLowercase{\textit{et al.}}: Bare Demo of IEEEtran.cls for IEEE Journals}
%% The only time the second header will appear is for the odd numbered pages
%% after the title page when using the twoside option.
%%
%% *** Note that you probably will NOT want to include the author's ***
%% *** name in the headers of peer review papers.                   ***
%% You can use \ifCLASSOPTIONpeerreview for conditional compilation here if



\maketitle

\begin{abstract}

Fine-grained object categorization aims for distinguishing objects of subordinate categories that belong to the same entry-level object category. It is a rapidly developing subfield in multimedia content analysis. The task is challenging due to the facts that (1) training images with ground-truth labels are difficult to obtain, and (2) variations among different subordinate categories are subtle. It is well established that characterizing features of different subordinate categories are located on local parts of object instances. However, manually annotating object parts requires expertise, which is also difficult to generalize to new fine-grained categorization tasks. In this work, we propose a Weakly Supervised Part Detection Network (PartNet) that is able to detect discriminative local parts for the use of fine-grained categorization. A vanilla PartNet builds on top of a base subnetwork two parallel streams of upper network layers, which respectively compute scores of classification probabilities (over subordinate categories) and detection probabilities (over a specified number of discriminative part detectors) for local regions of interest (RoIs). The image-level prediction is obtained by aggregating element-wise products of these region-level probabilities, and meanwhile diverse part detectors can be learned in an end-to-end fashion under the image-level supervision. To generate a diverse set of RoIs as inputs of PartNet, we propose a simple Discretized Part Proposals module (DPP) that directly targets for proposing candidates of discriminative local parts, with no bridging via object-level proposals. Experiments on benchmark datasets of CUB-200-2011, Oxford Flower 102 and Oxford-IIIT Pet show the efficacy of our proposed method for both discriminative part detection and fine-grained categorization. In particular, we achieve the new state-of-the-art performance on CUB-200-2011 and Oxford-IIIT Pet datasets when ground-truth part annotations are not available.

\end{abstract}

\begin{IEEEkeywords}
Fine-grained object categorization, part proposal, weakly supervised learning
\end{IEEEkeywords}

% \fi
\IEEEpeerreviewmaketitle

\section{Introduction}

\begin{figure}[!htb]
	\begin{center}
			\centering
			\includegraphics[width=0.98\linewidth]{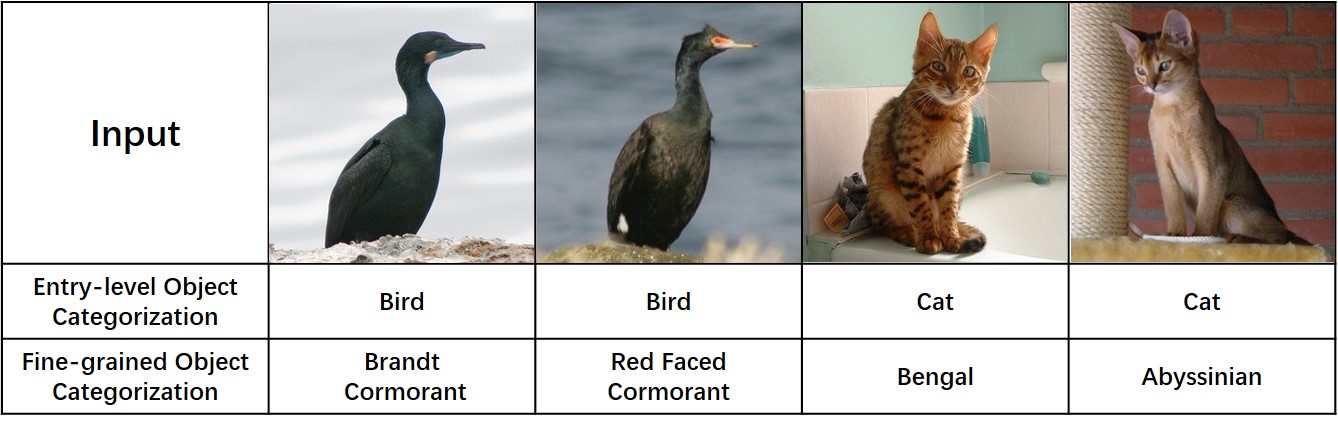}
	\end{center}
	\caption{Entry-level object categorization \emph{versus} fine-grained object categorization. In entry-level object categorization, we only need to distinguish the first two images of ``Bird'' from the last two images of ``Cat''. In fine-grained object categorization, the subcategories belonging to the same entry-level one should be further differentiated.}
	\label{fig:entry_fine}
\end{figure}

\IEEEPARstart{F}{ine}-grained object categorization aims for distinguishing objects of subordinate categories that belong to the same entry-level object category, e.g., various species of birds \cite{cub2002010, cub2002011, birdsnap}, pets \cite{stanforddog, pets}, or flowers \cite{flowers}. The difference between entry-level and fine-grained object categorization is shown in Figure \ref{fig:entry_fine}. Owing to its importance in a wide variety of applications, e.g., multimedia information retrieval \cite{fg_image_search, clothing_retrieval, landmark_classification}, e-commerce \cite{e_commerce} and rich image captioning \cite{rich_cap1, rich_cap2}, fine-grained object categorization has attracted widespread attention from multimedia community. However, the fine-grained categorization tasks are challenging because the variations among different subordinate object categories are subtle, which are often overwhelmed by those caused by arbitrary poses, viewpoint change, and/or occlusion. It is also difficult to obtain and label a large number of training images of subordinate object categories. Consequently, the performance of fine-grained categorization lies behind that of generic object recognition.

It is well known that characterizing features of different subordinate object categories are located at some local parts of object instances (e.g., the head and body of bird as illustrated in Figure \ref{fig:detection_scores}).
Correspondingly, many fine-grained categorization datasets provide ground-truth part annotations \cite{cub2002011}, \cite{birdsnap}. Existing methods \cite{partstacked, partbased, finegrainedpose, spdacnn}, \cite{task-driven} use these part annotations to train detection models that can detect from input images the most discriminative parts for the use of fine-grained categorization. However, manually annotating object parts requires expertise, which is also difficult to generalize to fine-grained categorization tasks of new entry-level object categories. To get relief from manual part annotations, a number of recent methods \cite{dvan, lookcloser, theapplication, weaklysupervised, opaddl, macnn}, \cite{pdfs, picking_journal}, \cite{surrogate-part-features} are proposed that aim for mining and leveraging discriminative local parts using image-level category labels only. Weakly supervised learning \cite{theapplication, weaklysupervised, opaddl}, \cite{pdfs, picking_journal} and attention mechanism in deep networks \cite{dvan, lookcloser, macnn}, \cite{surrogate-part-features} are the two main workhorses to achieve such a goal. Given region proposals \cite{selectivesearch}, weakly supervised learning based methods use a separate stage of region clustering \cite{theapplication, weaklysupervised, opaddl} or region mining \cite{pdfs, picking_journal} to learn part detectors, which is suboptimal for the final task of fine-grained categorization. Attention-based methods \cite{dvan, lookcloser}, \cite{surrogate-part-features} overcome such a limitation by automatically identifying and using salient/discriminative pixels and regions in an end-to-end fashion. However, they seem to have the weakness that a diverse set of discriminative parts are difficult to obtain, which restricts their practical performance.

To address the above limitations, we propose in this work a novel fine-grained categorization architecture called Weakly Supervised Part Detection Network (PartNet). A vanilla PartNet builds on top of a base convolutional (conv) subnetwork two parallel streams of upper network layers: given proposed regions of interest (RoIs), the \textit{classification stream} performs region-level differentiation over subordinate object categories and outputs classification probabilities; the \textit{detection stream} learns a specified number of part detectors that assign association probabilities of these RoIs with each of the learned detectors; the final image-level prediction is obtained by aggregating element-wise products of region-level probabilities of the two streams. PartNet training uses image-level supervision that enables the detection stream to achieve end-to-end learning of diverse part detectors in a weakly supervised manner.

Our proposed PartNet requires proposals of RoIs as inputs of the classification and detection streams. Existing fine-grained categorization works \cite{theapplication}, \cite{weaklysupervised}, \cite{opaddl}, \cite{pdfs, picking_journal} either directly use regions provided by off-the-shelf object proposal methods such as Selective Search (SS) \cite{selectivesearch}, or segment regular sub-regions from object proposals. However, criteria of object proposal methods are designed for region completeness of object instances, with no mechanism of proposing discriminative local parts; segmenting regular sub-regions from object proposals is an indirect approach to discriminative part proposals. Inspired by the discretization of proposal space in Region Proposal Networks (RPN) \cite{faster-rcnn}, we introduce in this work a simple but effective Discretized Part Proposals module (DPP). Our part proposals are anchored at salient locations in individual spatial cells of feature maps, where activation values are of higher magnitude. Correspondingly, candidates of discriminative local parts can be proposed independently of spatial locations of (possibly false positive) object instances. Experiments on benchmark fine-grained object categorization datasets show the efficacy of the proposed method. We summarize major contributions of this work as follows.
\begin{itemize}
\item We introduce in this paper a novel fine-grained categorization architecture called PartNet (cf. Section \ref{SecPartNet}). By using parallel classification and detection streams that process RoI features and aggregating their region-level scores, the proposed PartNet achieves end-to-end learning of diverse part detectors in a weakly supervised manner.

\item Existing region proposal methods focus on completeness of object-level regions, which is not directly relevant to proposing candidates of discriminative local parts. We introduce in this work a simple but effective DPP (cf. Section \ref{dpp}), which supports the success of PartNet for fine-grained categorization and could also be useful to other tasks that rely on discriminative local features.

\item We present a few variants of PartNet including (1) PartNet with the higher resolution of feature maps and (2) PartNet with orthogonal weight matrix in the classification stream (cf. Section \ref{SecPartNetVariants}). Experiments on the benchmark datasets of CUB-200-2011, Oxford Flower 102 and Oxford-IIIT Pet show that our proposed PartNet and its variants are effective for both discriminative part detection and fine-grained categorization. In particular, we achieve the new state-of-the-art performance on the CUB-200-2011 and Oxford-IIIT Pet datasets when ground-truth part annotations are not available (cf. Section \ref{SecExp}).
%we obtain excellent result, which is comparable with the state-of-the-art performance, on the CUB-200-2011 dataset when ground-truth part annotations are not available (cf. Section \ref{SecExp}). % we achieve the new state-of-the-art performance on the CUB-200-2011 dataset when ground-truth part annotations are not available (cf. Section \ref{SecExp}).}
\end{itemize}

\section{Related Works}
\label{SecLiterature}

In this section, we first present a brief review of fine-grained object categorization methods. We discuss how discriminative parts among fine-grained categories are essential for this task, with special focus on those methods that do not rely on ground-truth part annotations. We also discuss methods of object/part proposal and weakly supervised object detection, which are the techniques closely related to our proposed method.

\subsection{Part-Aware Fine-Grained Object Categorization}

Since the introduction of fine-grained categorization tasks, researchers realize that extracting features from discriminative local parts is essential to the success of the task. For example, \cite{hsnet} sequentially searches discriminative parts by unifying heuristic function and successor function via a Long Short-Term Memory network (LSTM). The heuristic function evaluates the informativeness of the proposed bounding boxes and the successor function predicts the offsets to the discriminative proposals of the proposed boxes. All the detected image parts are fused for fine-grained recognition. Jointly optimizing the fine-grained classification loss and the Euclidean distances between the proposed part proposals and the ground-truth part proposals, state-of-the-art result is obtained on the benchmark CUB-200-2011 dataset \cite{cub2002011}. To get relief from manual part annotations, recent efforts resort to weakly supervised learning \cite{theapplication,weaklysupervised,opaddl}, \cite{pdfs, picking_journal} and/or attention mechanism in deep networks \cite{dvan,lookcloser,macnn}, \cite{surrogate-part-features}, in order to either implicitly make use of information of salient parts \cite{dvan}, \cite{surrogate-part-features}, or explicitly identify discriminative local parts based on image-level category labels only \cite{lookcloser,weaklysupervised,theapplication,opaddl,macnn}, \cite{pdfs, picking_journal}. We briefly review some of these representative methods as follows.

Based on off-the-shelf object proposal methods (e.g., SS \cite{selectivesearch}), multi-scale part proposals are generated in \cite{weaklysupervised} at regular spatial grids of object proposals. These part proposals are then clustered from which useful ones are selected, in a weakly supervised manner, by measuring their importance scores for fine-grained categorization. Xiao {\it et al.} \cite{theapplication} also use image-level supervision and patch clustering to identify discriminative parts from patch proposals: a classifier of the entry-level category is first trained and used to filter out background patches; spectral clustering is then applied to the remaining patches to learn part detectors (e.g., cluster centers), which are further used to select discriminative parts from patch proposals; final classification is performed using features of the detected parts. Image-level supervision and object-part spatial constraint are applied to select the discriminative part proposals in \cite{opaddl}, and then neural clustering clusters selected proposals into semantic parts: a pre-trained entry level classifier is fine-tuned on target data and used to filter out noisy patches; object-level bounding boxes are obtained by class activation mapping (CAM) \cite{zhou2016detection} and used to further refine the selected proposals; part detectors, which are obtained by performing clustering on the neurons of a middle layer in the classification model, cluster selected proposals into diverse semantic parts.
Zhang {\emph{et al.}} \cite{pdfs,picking_journal} learn initial part detectors from distinctive region proposals by measuring activation outputs of network neurons; the detectors are refined via iterative alternation between new distinctive sample mining and part model retraining; neural activations are pooled into the final representation via a spatially weighted combination of Fisher Vectors coding, which considers the importance of each activation.
 % Object-level attention model and part-level attention model are proposed in \cite{opaddl}. Object-level attention model consists of patch filters and saliency extraction: a pre-trained entry-level category classifier is fine-tuned on target data and used to filter out noisy patches; saliency map of the object is obtained by the method of CAM \cite{zhou2016detection}. Part-level attention model is composed of object-part spatial constraint model and part alignment: the object-part spatial constraint model makes the selected parts share large overlap with the object but small overlap with each other; the part alignment performs clustering on the neurons of a middle layer in classifier model to cluster selected patches into diverse semantic parts.
In \cite{lookcloser}, multi-scale attention mechanism is employed into classification networks in order to guide deep feature learning to focus on discriminative (species-specific) regions, where starting from the full image, a hierarchy of three-level region scales are gradually attended, and their features are extracted for classification. Fine-grained categorization is obtained by integrating the information of three scale regions. In \cite{dvan}, a diversified LSTM based attention model is proposed that aims to learn a diverse set of discriminative region attentions, so that classification among fine-grained categories can rely more on features of these attended regions. In \cite{macnn}, multiple part attentions are generated by clustering, weighting from spatially-correlated convolutional channels. Part-level patches of each part and object-level images are taken as input to train individual part-CNN. The features of each part and object image of the part-CNNs are concatenated together for final classification. In \cite{surrogate-part-features}, activation values of feature maps are defined as assignment strengths for surrogate parts, and the part-level features are generated within the Bag-of-Words framework. Multi-scale and multi-position part features are obtained with the scale pooling and sub-region partition schemes on the feature maps respectively. The final image prediction is the product of the global image prediction and the part-level prediction achieved by averaging the parts' features.

Attention-based methods have the nice property that salient/discriminative pixels and regions can be automatically learned and attended in an end-to-end fashion. However, they seem to have the weakness that a diverse set of discriminative parts are difficult to obtain. \footnote{Even if automatic detection of salient regions is enabled by attention-based methods, it seems that explicit region proposals (e.g., via multi-scale proposals at regular spatial grids) always help. In fact, regions of varying sizes are cropped in \cite{dvan} at different locations of the original image in order to provide more diversified attention canvas.} For example, only one (but multi-scale) part is attended in \cite{lookcloser}; consequently, other potentially discriminative parts are ignored in classification. On another hand, existing methods based on explicit region proposals \cite{weaklysupervised,theapplication} use a separate stage of region clustering to obtain part detectors, which is suboptimal for the final task of fine-grained categorization. While our proposed PartNet also relies on explicit region proposals, we employ in the upper network parallel streams of classification and detection, which simultaneously achieve discriminative part detection and fine-grained categorization. The detection stream also enables learning of diverse part detectors. Superior results on the benchmark datasets \cite{cub2002011}, \cite{flowers} show the efficacy of our proposed method.

\subsection{Weakly Supervised Object Detection/Localization}

Weakly supervised object detection/localization aims to learn object detectors using only image-level category labels, i.e., ground-truth object annotations (e.g., object bounding boxes) are not required. Simple extensions of such techniques could be useful for fine-grained categorization by learning part detectors in a weakly supervised manner. There are many weakly supervised object detection/localization methods proposed in the literature, among which CNN based methods show great promise recently \cite{zhou2016detection,wsddn,contex,midn,weak_2_strong}. This may be due to the fact that CNNs have remarkable localization ability despite being trained on image-level labels \cite{zhou2016detection}. We particularly mention here a model of Weakly Supervised Deep Detection Networks (WSDDN) \cite{wsddn}. It introduces a two-stream network architecture where the classification stream differentiates each object proposal among different object categories, and the detection stream ranks for each category all the object proposals. Scores from the two streams are aggregated via element-wise product, which are finally used for image-level supervision. Our proposed PartNet is inspired by \cite{wsddn}. Instead of ranking object proposals for each category in the detection stream, we rely on part proposals and learn multiple part detectors that altogether contribute to the classification of fine-grained categories.

\subsection{Generation of Object/Part Proposals}

Both our proposed PartNet and other part-aware fine-grained categorization methods \cite{weaklysupervised,theapplication,opaddl} rely on proposals of local object regions/parts. Part proposals differ from the established object proposal techniques \cite{selectivesearch,edgebox} in that salient part locations and part boundaries are less clearly defined. Consequently, it is less obvious to extend existing object proposal techniques for a good part proposal method. Nevertheless, existing efforts either directly use object proposal methods for use of part proposals, e.g., SS \cite{selectivesearch} is used in \cite{theapplication}, \cite{opaddl}, or simply use sub-regions of object proposals as part proposals \cite{weaklysupervised}. In this work, we propose a simple DPP that borrows the idea of spatial space discretization from RPN \cite{faster-rcnn}. Our part proposals are anchored at discriminative locations of feature maps, which are obtained by training using image-level category labels only. Comparative studies with SS \cite{selectivesearch} show the efficacy of our proposed DPP.

\begin{figure*}[t]
	\begin{center}
		\includegraphics[width=0.8\linewidth]{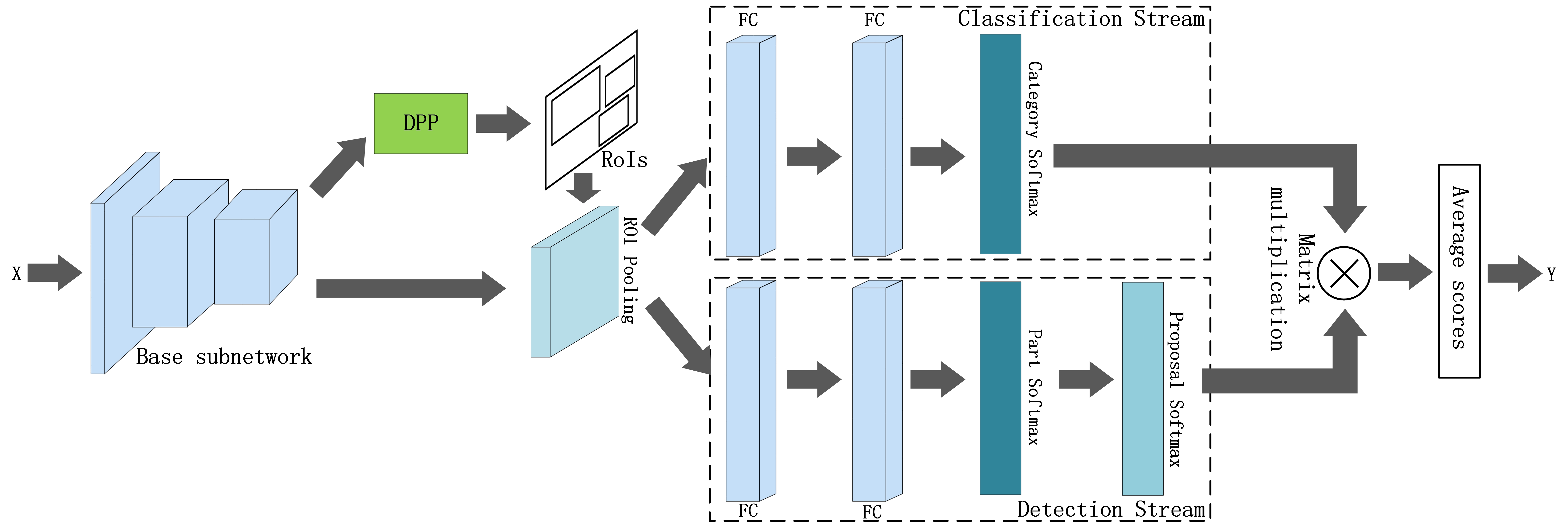}
	\end{center}
	\caption{Framework of the PartNet.
%Weakly Supervised Part Detection Network (PartNet).
The base subnetwork represents the convolutional layers that are pre-trained on ImageNet dataset firstly and then fine-tuned on the fine-grained training data. The DPP represents our proposed module of Discretized Part Proposals (cf. Section \ref{dpp}) for generating RoIs. The classification stream differentiates region-level proposals over subordinate object categories, while the detection stream assigns association probabilities of those proposals with part detectors. Region-level probabilities of the two streams are combined with matrix multiplication. The image-level classification is obtained by averaging the classification probabilities of different detected parts. The different softmax layers are detailed in Section \ref{SecPartNet}.}
	\label{fig:the pafgn model structure}
\end{figure*}

\section{The Proposed Method}
\label{proposed method}
In this section, we present in details our proposed PartNet, which is empowered by a simple but effective DPP module.
We also introduce a few variants of PartNet that altogether contribute to an effective solution to part-aware fine-grained categorization.

%Discretized Part Proposals module (DPP).
%Weakly Supervised Part Detection Network (PartNet),

\subsection{Weakly Supervised Part Detection Network}
\label{SecPartNet}

\begin{figure*}

    \subfigure[Values of $\tilde{s}_{det}^{ij}$ for the example region proposals]{
    \label{fig:local_part}
    \noindent
    \begin{minipage}{0.79\textwidth}
    \includegraphics[width=\textwidth]{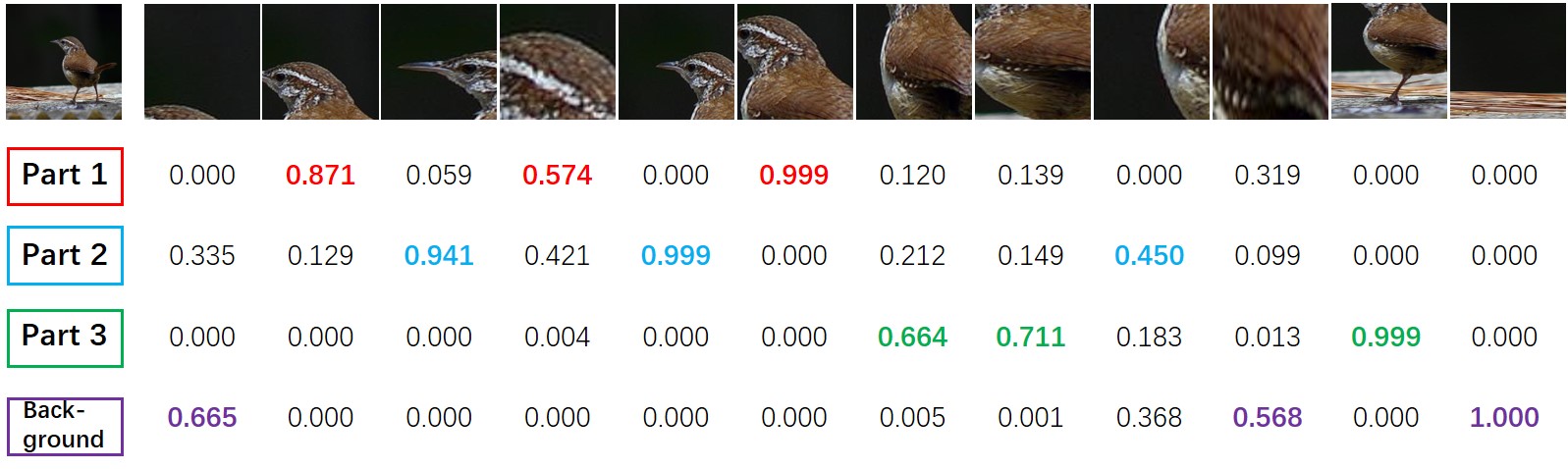}
    \end{minipage}
    }
    \hfill
    \subfigure[Top-scored parts]{
    \begin{minipage}{0.165\textwidth}
      \includegraphics[width=\textwidth]{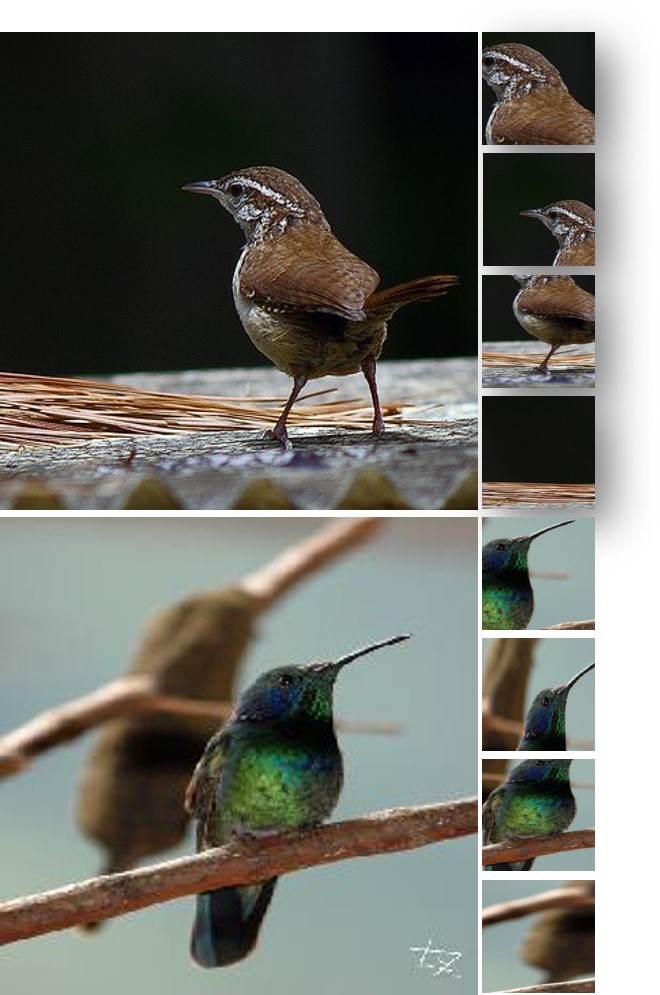}
    \end{minipage}
    \label{fig:v_two_birds}
    }

	\caption{(a) Visualization of the detection scores $\tilde{s}_{det}^{ij}$ of equation (2) when applying PartNet to the CUB-200-2011 dataset, where score precision is rounded to the level of $10^{-3}$. The first row shows the input image (left) and example region proposals (right) generated by the DPP module. The second, third, fourth, and bottom rows respectively present the scores of three part detectors and the background detector for each proposal. (b) Visualization of input images and their respective detected (top-scored) local parts. Please refer to Figure \ref{fig:part detector} for more examples of detected local parts.}
	 \label{fig:detection_scores}
\end{figure*}

As discussed in Section \ref{SecLiterature}, it is well established that identification of discriminative local parts is essential for fine-grained categorization. In this work, we design a novel architecture called PartNet (cf. Figure \ref{fig:the pafgn model structure} for an illustration), which explicitly learns part detectors using image-level category labels only. Examples of our detected local parts from the fine-grained categorization datasets are also shown in Figure \ref{fig:detection_scores} and Figure \ref{fig:part detector}.

A vanilla PartNet uses conv layers as its base subnetwork. Assume the final conv layer of the base subnetwork outputs $N$ feature maps. Using our proposed DPP (cf. Section \ref{dpp}), a number $R$ of local regions of those feature maps are proposed that give RoIs on the input image. These RoIs are of varying sizes, and we use RoI pooling \cite{fastrcnn} to produce features of the fixed size  $m\times m$, which, after vectorization, gives a feature vector $\mathbf{f}_{RoI} \in \mathbb{R}^{Nm^2}$ for each proposed RoI. We use two parallel streams of fully connected (FC) layers on top of the base subnetwork to further process, in a batch mode, these RoI features $\{ \mathbf{f}_{RoI} \}$. Assume there are $C$ fine-grained object categories in the considered task. The \textit{classification stream} performs differentiation of the proposed RoIs among these categories. The \textit{detection stream} learns a specified number $P$ of patterns of parts (i.e., part detectors) that can identify from the proposed RoIs the most effective ones for fine-grained categorization. The two streams output part-level scores of classification/detection probabilities, which are then aggregated and used for image-level training or inference. We present component-wise specifics of our proposed PartNet as follows.

\vspace{0.2cm}
\subsubsection{The classification stream}

As shown in Figure \ref{fig:the pafgn model structure}, we use two  consecutive FC layers (with ReLUs) to differentiate each of the RoI feature vectors $\{ \mathbf{f}_{RoI}^i \}_{i=1}^{R}$ into fine-grained categories. Since some of the proposed RoIs are on the background, which are in fact common in different fine-grained categories, we introduce an additional output neuron in the second FC layer that corresponds to the \textit{background category}. The second FC layer thus outputs a matrix $\mathbf{X}_{cls}$ of the size $(C+1)\times R$.  A softmax operator, termed as ``category softmax'', is then followed to make $\mathbf{X}_{cls}$ as a score matrix $\mathbf{S}_{cls} \in \mathbb{R}^{(C+1)\times R}$ of classification probabilities. Elements of $\mathbf{S}_{cls}$ are computed as
\begin{eqnarray}\label{EqnClsSoftMax}
s_{cls}^{ij} = \frac{ e^{x_{cls}^{ij}} }{ \sum_{c=1}^{C+1}e^{x_{cls}^{cj}}  } ,
\end{eqnarray}
where $x_{cls}^{ij}$ is an entry of $\mathbf{X}_{cls}$, and $i$ and $j$ index the categories and RoI features respectively.

\vspace{0.2cm}
\subsubsection{The detection stream}
\label{SecDetectStream}

The detection stream aims for learning a specified number $P$ of part detectors that detect from (the proposed RoIs of) the input image local parts that are most useful/discriminative for fine-grained categorization. To this end, we use two consecutive FC layers (with ReLUs) to process RoI features $\{ \mathbf{f}_{RoI}^i \}_{i=1}^{R}$. To model those local parts that are either on the background or less discriminative among fine-grained categories, we use $P+1$ output neurons in the second FC layer. Outputs of the second FC layer are denoted as $\mathbf{X}_{det} \in \mathbb{R}^{(P+1)\times R}$. FC layers themselves barely give the detection stream the ability to learn distinctive and semantically meaningful part detectors.
% \textcolor {red} { \sout { Inspired by \cite{},}}
We use a softmax operator, termed as ``part softmax'', immediately following the second FC layer, which gives the output matrix $\widetilde{\mathbf{S}}_{det} \in \mathbb{R}^{(P+1)\times R}$. Elements of $\widetilde{\mathbf{S}}_{det}$ are computed as
\begin{eqnarray}\label{EqnDetSoftMax1}
\tilde{s}_{det}^{ij} = \frac{ e^{x_{det}^{ij}} }{ \sum_{p=1}^{P+1}e^{x_{det}^{pj}}  } ,
\end{eqnarray}
where $x_{det}^{ij}$ is an entry of $\mathbf{X}_{det}$, and $i$ and $j$ index the part detectors and RoI features respectively. While there are no ground-truth part annotations available, learning part detectors is made possible by the use of part softmax (cf. Eq. (\ref{EqnDetSoftMax1})): in the forward pass, each proposed RoI is associated with one of the $P+1$ output neurons of the second FC layer by scaling up the corresponding score toward the value of $1$ while suppressing others; this is reinforced in the backward pass and consequently, patterns of discriminative parts are learned as parameters of FC layers in a weakly supervised and locally optimal manner.
To give an intuition on what we have learned for part detectors, we illustrate in Figure \ref{fig:detection_scores} examples of the proposed RoIs for an input image, where scores in each column are the ones computed from (2) for each RoI, and rows of different colors correspond to individual part detectors, including the background one. Figure \ref{fig:detection_scores}-(a) shows that these RoIs are differentiated and associated with different part detectors, and those associated with the same one have the similar visual appearance, suggesting that individual part detectors are trained to characterize patterns of local distinctiveness. Figure \ref{fig:detection_scores}-(b) also shows that when applying these learned part detectors to images of different categories, they detect local regions that have the potential of fine-grained discrimination.

Then, we use a second softmax operator, termed as ``proposal softmax'', on $\widetilde{\mathbf{S}}_{det}$ to rank their associations with each of the $P$ part detectors. This produces $\mathbf{S}_{det} \in \mathbb{R}^{(P+1)\times R}$ whose elements are compute as
\begin{eqnarray}\label{EqnDetSoftMax2}
s_{det}^{ij} = \frac{ e^{\tilde{s}^{ij}} }{ \sum_{r=1}^R e^{\tilde{s}^{ir}}  } .
\end{eqnarray}
The second softmax also serves as a normalization layer that normalizes RoI scores associated with each part detector (i.e., each row of $\widetilde{\mathbf{S}}_{det}$), so that the resulting $\mathbf{S}_{det}$ can be better used for score aggregation with those of the classification stream, as explained shortly. In this work, we by default set the number $P$ of part detectors as $3$. We also investigate the effects of different values of $P$ on fine-grained categorization (cf. Section \ref{ExpDetection}).
%While there are no ground-truth part annotations available, learning part detectors is made possible by the use of the two softmax operations. In the forward pass of Eq. \ref{EqnDetSoftMax1}, each proposed RoI is associated with one of the $P+1$ output neurons of the second FC layer by scaling up the corresponding score toward the value of $1$, while suppressing others, which is reinforced in the backward pass. The Eq. \ref{EqnDetSoftMax2} also serves as a normalization layer that normalizes RoI scores associated with each part detector (i.e., each row of $\widetilde{\mathbf{S}}_{det}$), so that the resulting $\mathbf{S}_{det}$ can be better used for score aggregation with those of the classification stream, as explained shortly. Consequently, patterns of discriminative parts are learned as parameters of FC layers in a weakly supervised and locally optimal manner.

\vspace{0.2cm}	
\subsubsection{Aggregation of classification and detection scores for image-level supervision/inference} \label{Sec_aggregate_two_branches}

\begin{figure}[t]
\begin{center}
\includegraphics[width=0.8\linewidth]{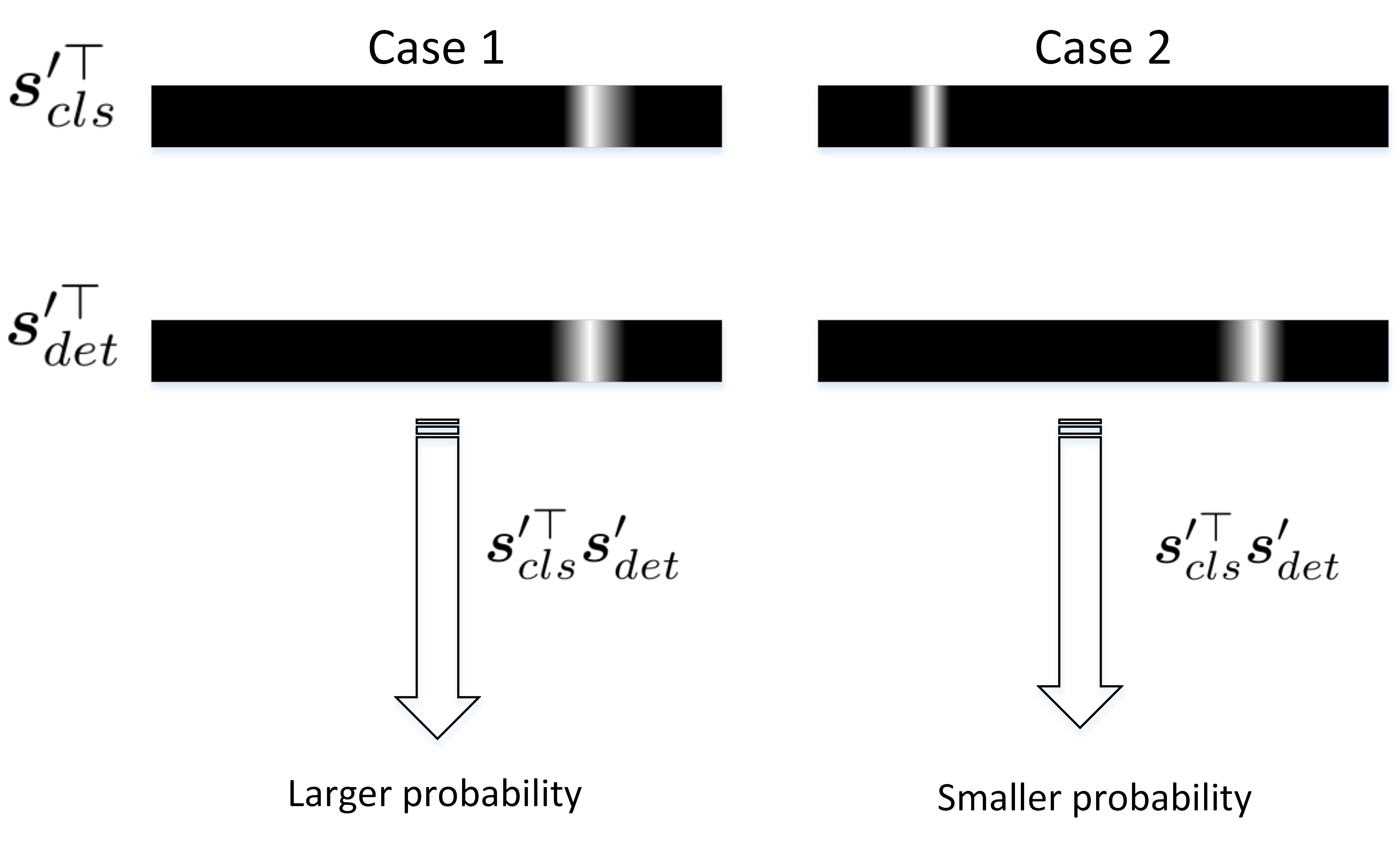}
\end{center}
\caption{In case 1, the RoIs that have larger values in $\mathbf{s}_{det}'$ also have larger values in $\mathbf{s}_{cls}'$, and the classification probability is larger. Otherwise, the classification probability is smaller in case 2. In order to achieve accurate classification, the RoIs, that have larger values in the right category of $\mathbf{S}_{cls}'$, should consistently have larger values in $\mathbf{S}_{det}'$.}
\label{fig:demo of combine two branch}
\end{figure}

The classification and detection streams output score matrices $\mathbf{S}_{cls}$ and $\mathbf{S}_{det}$ respectively for all proposals. To use them for image-level supervision or inference, we first remove from $\mathbf{S}_{cls}$ the last row that represents the probabilities of RoIs' belonging to the background category, and also remove from $\mathbf{S}_{det}$ the last row that contains scores of RoIs associated with the background/irrelevant part detector, resulting in reduced matrices $\mathbf{S}_{cls}' \in \mathbb{R}^{C\times R}$ and $\mathbf{S}_{det}' \in \mathbb{R}^{P\times R}$ respectively. Suppose an input image is of the $c^{th}$ fine-grained category. We denote the $c^{th}$ row of $\mathbf{S}_{cls}'$ as $\mathbf{s}_{cls}'^{\top} \in \mathbb{R}^{R}$ that contains the probabilities that the $R$ RoIs are classified as the $c^{th}$ category. We similarly denote the $p^{th}$ row of $\mathbf{S}_{det}'$ as $\mathbf{s}_{det}'^{\top} \in \mathbb{R}^{R}$ that contains the probabilities that the $R$ RoIs are detected as instances of the $p^{th}$ discriminative parts. Discriminative part detection requires that RoIs that have larger values in $\mathbf{s}_{det}'$ (i.e., the detected instances of the $p^{th}$ part) should consistently have larger values in $\mathbf{s}_{cls}'$ (cf. Figure \ref{fig:demo of combine two branch} for an illustration).  We thus choose to use $\mathbf{s}_{cls}'^{\top}\mathbf{s}_{det}'$ as a measure of part-level classification confidence. Write compactly in a matrix form we have $\mathbf{S}_{cls}'\mathbf{S}_{det}'^{\top} \in \mathbb{R}^{C\times P}$, each row of which contains the probabilities that the detected discriminative parts are of a certain fine-grained category. We then average the part-level probabilities to form the image-level classification representation $\mathbf{y} \in \mathbb{R}^C$ of the input image as
\begin{eqnarray}\label{EqnScoreAggreg}
\mathbf{y} = \frac{1}{P}\mathbf{S}_{cls}'\mathbf{S}_{det}'^{\top} \mathbf{1}_P ,
\end{eqnarray}
where $\mathbf{1}_P$ denotes a $P$-dimensional vector with all values of $1$. Note that as mentioned in Section \ref{SecDetectStream}, the proposal softmax (cf. Eq. (\ref{EqnDetSoftMax2})) in the detection stream serves as a normalization layer that ensures each entry value of $\mathbf{S}_{cls}'\mathbf{S}_{det}'^{\top}$ is in the range $[0, 1]$. Consequently, the computed $\mathbf{y}$ in Eq. (\ref{EqnScoreAggreg}) can be considered as image-level classification probabilities.

We use the result of Eq. (\ref{EqnScoreAggreg}) as the inference of a PartNet for an input image. To train the PartNet, assume a set of $M$ training images are given, each of which has its one-hot vector form of ground-truth category label as $\mathbf{g} \in \mathbb{R}^C$. Denote parameters of the PartNet collectively as a vector $\theta$, we use the following loss of binary cross-entropy to train the network
\begin{eqnarray}\label{EqnNetTrainLoss}
\frac{\lambda}{2} \| \theta \|_2^2  - \sum_{i=1}^M\sum_{j=1}^C g_{ij} \log y_{ij}(\theta) + (1 - g_{ij})\log(1-y_{ij}(\theta)) ,
\end{eqnarray}
where $\mathbf{g}_i$ and $\mathbf{y}_i$ are respectively the ground truth label and inference result for the $i^{th}$ training sample, and $g_{ij}$ and $y_{ij}$ are their $j^{th}$ entries. We optimize Eq. (\ref{EqnNetTrainLoss}) using Stochastic Gradient Descent (SGD) with momentum.

\begin{figure*}[t]%%
	\begin{center}
		\includegraphics[width=0.8\linewidth]{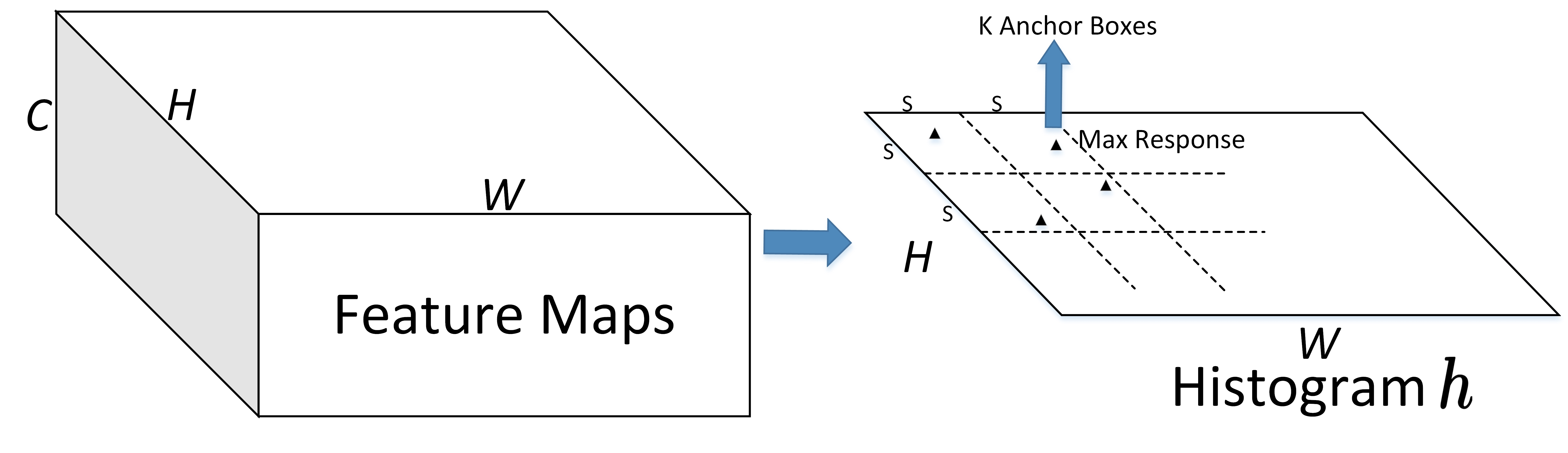}
	\end{center}
	\caption{The framework of the DPP module.
%Discretized Part Proposals module (DPP).
The DPP takes as input the feature maps that are generated by the last convolutional layer. A histogram, which is obtained by counting the occurrence of channel-wise peak value for each of the spatial locations, is firstly generated to measure the degrees of discrimination for different spatial locations. All the spatial locations are divided into $S\times S$ non-overlapping cells, and $K$ anchor boxes of varying sizes and aspect ratios are proposed anchored at the location corresponding to the max count of the histogram for each cell.}
	\label{fig:PRPN}
\end{figure*}

\subsection{Discretized Part Proposals in Spatial Cells of Feature Maps}
\label{dpp}

The PartNet presented in Section \ref{SecPartNet} needs proposals of RoIs that specify local regions of input images for classification and detection streams to work on. Existing part proposal methods \cite{weaklysupervised}, \cite{opaddl}, \cite{theapplication} either directly use regions provided by object proposal method \cite{selectivesearch}, or use their regular sub-regions. However, object proposal methods use criteria that focus on region completeness of object instances and are not effective by design for proposing candidates of discriminative parts. Segmenting regular sub-regions from object proposals can help, but it is not a direct approach to discriminative part proposals. In this work, we propose a simple DPP method
%Discretized Part Proposals (DPP) method
towards this goal. Our method is inspired by the discretization of proposal space in RPN \cite{faster-rcnn}; but we do not have a training process since ground-truth part annotations are not available.

It is well known that CNNs have a remarkable localization ability despite being trained using image-level labels \cite{zhou2016detection}, and ideally the discriminative parts should locate at positions of feature maps that have larger feature values. A similar idea is also adopted in \cite{surrogate-part-features} where the values of feature maps are defined as assignment strengths for surrogate parts. We thus opt to generate part proposals anchored at these positions directly. More specifically, given feature maps of the size $C \times W \times H$ that have $C$ channels, we calculate a histogram vector $\mathbf{h} \in \mathbb{R}^{WH}$ that counts for each of the $W \times H$ spatial locations the occurrence that channel-wise peak value is located at the current position, and use the obtained $\mathbf{h}$ to identify discriminative spatial locations. The location of the peak value for each channel is also used in \cite{macnn}. Counts in the histogram $\mathbf{h}$ measure the degrees of discrimination for different spatial locations, and part proposals are anchored at those with more counts.

To make part proposals spatially spread over the feature maps, we regularly divide the $W \times H$ spatial locations into $S\times S$ non-overlapping cells (e.g., $S$ = 4), which produces the corresponding sub-vectors from the histogram vector $\mathbf{h}$. We use spatial locations corresponding to the max count of each histogram sub-vector as our anchors of part proposals. For each anchor position, we define $K$ anchor boxes of varying sizes and aspect ratios \cite{faster-rcnn}. We by default set $K = 28$ in our experiments, and Table \ref{anchor size} gives its box sizes and aspect ratios. The influence of using different $K$ values is also investigated in Section \ref{Exp:dpp}.

\begin{table}
       \caption{The Specified Sizes and Aspect Ratios When We Use $K = 28$ Anchor Boxes for Each Anchor Position on the Feature Maps. }\label{anchor size}
       \centering
       \setlength{\tabcolsep}{1.5mm}{
       \begin{tabular}{|c|c|c|c|c|c|c|c|c|c|c|}
         \hline
         Anchor sizes & $3^2$ & $5^2$ & $7^2$ & $9^2$ & $11^2$ & $13^2$ & $15^2$ & $17^2$ & $19^2$ & $21^2$ \\
         \hline
         Aspect ratios & 1:1 & \multicolumn {9}{|c|} {1:1  \quad     1:2    \quad   2:1} \\
         \hline
       \end{tabular}}
\end{table}

\subsection{Other Variants}
\label{SecPartNetVariants}

In this section, we present two variants of PartNet in order to boost the performance on fine-grained categorization tasks.

%\noindent\textbf{Higher resolution of feature maps via dilated convolution} Variations among fine-grained categories are often subtle, regional, and imaged in finer details. However, finer details could disappear when feature maps are of lower resolution. To avoid this issue, we present a variant of the vanilla PartNet as follows. Taking as the base subnetwork a ResNet model \cite{resnet} pre-trained on the ImageNet (after removing its last layer of classifier), we replace its last stride-2 conv layer (i.e., conv5\_1 in ResNet-34) with a stride-1 one, and modify the subsequent conv layers via 2-dilated convolution layers \cite{dilated}; parameters of these conv layers keep unchanged. By this way, the resolution of feature maps is doubled without affecting the size of receptive field.
\noindent\textbf{Higher resolution of feature maps} Variations among fine-grained categories are often subtle, regional, and imaged in finer details. However, finer details could disappear when feature maps are of lower resolution. To avoid this issue, we present a variant of the vanilla PartNet as follows by modifying the base subnetwork structure. For the models that downsample feature maps via stride-2 conv layers (e.g., ResNet \cite{resnet}), its last layer of the classifier is removed firstly, then we replace its last stride-2 conv layer (i.e., conv5\_1 in ResNet-34) with a stride-1 one, and modify the subsequent conv layers via 2-dilated conv layers \cite{dilated}. For the models that downsample feature maps via stride-2 max pooling layers (e.g., VGGNet \cite{vgg}), the last stride-2 max pooling layer and the subsequent layers are removed. By this way, the resolution of the base subnetwork feature maps is doubled.
%For the models that downsample feature maps via stride-2 max pooling layers (e.g. VGGNet \cite{vgg}), the last stride-2 max pooling layer is replaced by a global average pooling layer and the subsequent FC layers are replaced by a FC layer, the number of output neurons of that is the same as the ImageNet categories. By this way, the resolution of the base subnetwork feature maps is doubled.

\noindent\textbf{Orthogonal weight matrix in the classification stream} Orthogonal weight matrices are observed to be helpful to propagate information in deep networks \cite{svb}.
%, and also to be effective to produce more discriminative layer of classifier \cite{AAA} {\color{red} (I remember you ever mentioned that there is a related work that used orthogonal weight matrix for classification?)}.
In this work, we present a variant of the vanilla PartNet that applies the technique of Singular Value Bounding (SVB) \cite{svb} to the second FC layer in the classification stream. We expect this variant to produce more discriminative scores of classification probabilities among different fine-grained categories.

\subsection{Final Prediction}
\label{Sec_Final_Prediction}
The proposed PartNet achieves fine-grained categorization by aggregating regional discrimination of detected individual parts. However, each of the individual parts may independently contribute to fine-grained categorization their own discrimination. The input image may also provide complementary holistic features. To utilize all these part-level and image-level discriminative information, we adopt a region zooming strategy as in \cite{macnn,lookcloser,partbased,opaddl}.

Specifically, given a trained PartNet, the $P$ part detectors of the detection stream respectively rank the $R$ region proposals generated by DPP, resulting in a score matrix $\mathbf{S}'_{det} \in \mathbb{R}^{P\times R}$ (cf. Section \ref{Sec_aggregate_two_branches}). Intuitively, if features of a proposal $i$ match pattern of a part $j$, then the $(j, i)$ entry of $\mathbf{S'}_{det}^{ji}$ would have a larger value, otherwise it would have a smaller one. We thus select for each of the $P$ part detector $M$ region proposals of top scores (e.g., $M = 50$), and use the selected regions to fine-tune the image-level model, resulting in $P$ part-level models. During testing, the top-1 region proposal for each part detector is selected and zoomed as the input of the corresponding part-level model. Our final prediction is made by averaging the classification probability of PartNet with those of its associated image-level and the $P$ part-level models. We term such an ensemble model as PartNet-Full.

\section{Experiments}
\label{SecExp}

In this section, we conduct fine-grained categorization experiments on the benchmark datasets of CUB-200-2011 \cite{cub2002011}, Oxford Flower 102 \cite{flowers} and Oxford-IIIT Pet \cite{pets}.
We present ablation studies to investigate the component-wise effectiveness of our proposed PartNet, its variants, and the DPP scheme, and also compare with the state of the art. We implement the proposed method on PyTorch and provide the codes at \url{https://github.com/YBZh/PartNet}.

\subsection{Datasets and implementation details} \label{implementation details}

\noindent\textbf{CUB-200-2011 \cite{cub2002011}}  The Caltech-UCSD Birds 200-2011 dataset is the most widely-used dataset for fine-grained categorization and contains 200 species of birds. It includes 5,994 images for training and 5,794 images for testing. For each image, one bounding box annotation and 15 keypoint annotations are given. We do not use these bounding box or keypoint annotations in our experiments.

\noindent\textbf{Oxford Flower 102 \cite{flowers}} Oxford Flower 102 contains 102 categories of flowers. There are 1,020 images for training, 1,020 images for validation, and 6,149 images for testing. We do not use the image segmentations provided in this dataset, and instead, we only use the category labels in our experiments.

\noindent\textbf{Oxford-IIIT Pet \cite{pets}} Oxford-IIIT Pet contains 37 pet subcategories, among which 12 are cat subcategories and 25 are dog subcategories. There are 3,680 images for training and 3,669 images for testing. We do not use the pixel level segmentation provided in this dataset, and instead, we only use the category labels in our experiments.

%\noindent\textbf{Stanford Dogs \cite{stanforddog}} The Stanford Dogs dataset contains 120 different types of dogs. There are 12,000 images for training and 8580 images for testing. The bounding boxes are provided in this dataset, however we only use the subcategory labels in our experiments.

%\noindent\textbf{Cars-196 \cite{stanfordcar}} It contains 16,185 images of 196 classes of cars. The dataset includes 8,144 images for training and 8,041 images for testing. Each image is annotated with a subcategory label and a object bounding box. In our experiments, only the subcategory labels are used.

%\noindent\textbf{FGVC-Aircraft \cite{aircraft}} There are 10,000 images of 100 classes of aircrafts. It includes 6667 images for training and 3333 images for testing.

\noindent\textbf{Baselines and implementation details} We first present the baseline models that we use to compare with our proposed methods. Given a 34-layer ResNet \cite{resnet} that is pre-trained on the ImageNet \cite{imagenet}, we modify its final FC layer of the classifier to make the number of its output neurons the same as that of the target fine-grained categories. This baseline model is termed as ResNet-34. To fairly compare the baseline with variants of our proposed PartNet, we also introduce an additional baseline of Dilated ResNet-34, which is obtained by modifying ResNet-34 as the way described in Section \ref{SecPartNetVariants}. The ResNet-34 and Dilated ResNet-34 based models are used in our ablation studies. To investigate the efficacy of our proposed contributions when comparing with many of the existing methods, we also construct our methods on the VGGNet \cite{vgg}. The VGGNet, where batch normalization \cite{bn} is used to improve network training, is pre-trained on the training images of ImageNet dataset \cite{imagenet} firstly and then we modify its structure as the way described in Section \ref{SecPartNetVariants}. Then a FC layer is followed as the fine-grained categories classifier. Those baseline models are fine-tuned on the target fine-grained categorization datasets and are termed as image-level models.
%We first modify the VGG-19 with batch normalization \cite{bn} as the way described in Section \ref{SecPartNetVariants}.
%\textcolor {red} {The VGGNet, where batch normalization \cite{bn} is used to improve network training, is pre-trained on the training images of ImageNet dataset \cite{imagenet} and then we modify its final FC layer of classifier to make the number of its output neurons the same as that of the target FGVC categories.} Those baseline models are fine-tuned on the target FGVC datasets and are referred as image-level models.

% {\color{red} CAN YOU COMPARE THE ABOVE REVISED PARAGRAPH WITH BELOW (YOUR PREVIOUS ONE) TO LEARN SOMETHING? --- We modify the pre-trained 34-layer and 101-layer residual network to ``larger resolution of feature maps'' variant following Section \ref{SecPartNetVariants}. Then we change the kernel size of the Global Average Pooling(GAP) layer to 28$\times$28 and change the output size of the final output layer to the number of the fine-grained categories. The modified models are termed as Dilated Resnet-34 and Dilated Resnet-101. We fine-tune the Dilated Resnet models and Resnet models on the training data of the fine-grained dataset as our baselines to verify the effectiveness of the PartNet. Cross Entropy Criterion is used to train the networks. }

Our proposed PartNet (and its variants) are constructed based on the above image-level models. Taking the ResNet-34 as an example, we build up the base subnetwork of PartNet by removing its layer of global average pooling and also the subsequent (final) layer of classifier; we then use an RoI pooling layer \cite{fastrcnn} whose inputs are formed by the part proposals generated by DPP, together with output feature maps of the base subnetwork; following the RoI pooling layer, a parallel pair of detection and classification streams are used that respectively produce scores of detection and classification probabilities; these scores are finally aggregated and used for image-level training or inference (cf. Section \ref{SecPartNet}). Figure \ref{fig:the pafgn model structure} gives an illustration.
      	
% {\color{red} CAN YOU COMPARE THE ABOVE REVISED PARAGRAPH WITH BELOW (YOUR PREVIOUS ONE) TO LEARN SOMETHING? --- Our PartNet is based on the baseline model that has been fine-tuned on the target FGVC task. We modify the baseline model following the illustrations in Section \ref{proposed method} to construct the PartNet: First, we cut off the GAP layer and the following layers in the baseline model.  Then a DPP module (cf. Section \ref{dpp}) is added following the last convolution layer (conv5\_3 in the residual network family) and a ROI-Pooling layer is built on the top of the DPP and last convolution layer. Second, we parallelly add the detection stream and classification stream following the ROI-Pooling layer, and the output of the two streams are combined with matrix multiplication (cf. Section \ref{SecPartNet}). Binary Cross Entropy Criterion (Eq. \ref{EqnNetTrainLoss}) is used to train the PartNet.}
      	
To train the above models, we use SGD with momentum: we set the weight decay as 1e-4 and momentum as 0.9; we train each model for 160 epochs with a batch size of 128; for parameters that are initialized from pre-trained models, we use a learning rate of 1e-3; for other parameters, we use an initial learning rate of 1e-1, which drops by a factor of 10 respectively after 80 and 120 epochs. For inputs of PartNet and image-level models, we pre-process each image by resizing its shorter size to 448 while keeping the aspect ratio unchanged. Then we crop a random $448\times 448$ region for the use of training (we also use the horizontal flip version of the cropped $448\times 448$ region for data augmentation) and a central $448\times 448$ region for the use of testing. The part-level models are obtained by fine-tuning the image-level model with the detected part proposals, which are rescaled to the size of $448\times 448$ as inputs. All our experiments are based on the above training settings.

Based on models constructed from VGGNet, we report the time to train the models and to label a new sample on Tesla M40 GPUs. Taking the CUB-200-2011 dataset as an example, it takes $19.4$, $66.8$ and $89.8$ GPU hours to train the image-level model, PartNet, and each part-level model respectively. Thus, the whole training time of our method is $355.6$ GPU hours, and the model training can be finished in $176$ GPU hours considering that the three part-level models can be trained in parallel. It takes $32$ ms, $71$ ms and $32$ ms to label a new sample by image-level model, PartNet, and each part-level model respectively. Thus the prediction of a new sample by the PartNet-Full can be finished in $103$ ms considering that predictions of the image-level model and PartNet can be made in parallel, and the same applies to predictions of the three part-level models. The number of parameters of PartNet ($58.63$ M) is about $41\%$ of those of the original VGGNet model (about $144$ M), confirming the efficiency of our proposed method.
\color {black}

\subsection{Ablation Studies on the Detection Stream} \label{ExpDetection}

The detection stream is the key component in PartNet. We evaluate its effectiveness on the CUB-200-2011 dataset using a PartNet constructed from Dilated ResNet-34.

The detection stream detects discriminative local parts essentially by learning to assign varying weights to different region proposals. To evaluate its effectiveness, we remove the detection stream of PartNet and correspondingly set scores of detection probabilities for different region proposals as being equal (i.e., setting elements of $\mathbf{S}_{det}'$ in Eq. (\ref{EqnScoreAggreg}) as $\frac{1}{R}$). We term such a model as Degenerate PartNet. Table \ref{detection_ablation} compares results of PartNet, Degenerate PartNet, and also the baseline Dilated ResNet-34, where region proposals are generated by SS \cite{selectivesearch}. Dilated ResNet-34 performs fine-grained categorization directly on the image level, and its result is worse than that of Degenerate PartNet, showing the benefit of performing fine-grained categorization on the region level. This is consistent with observations in \cite{theapplication, opaddl}. By learning and assigning varying weights to different region proposals, our proposed PartNet further improves the result. Discriminative local parts can also be detected from region proposals by ranking these weights, which will be presented shortly.

\begin{table}[htbp]
              \caption{Comparative Experiments on the CUB-200-2011 Dataset \cite{cub2002011} With or Without the Detection Stream in the PartNet, Which Is Constructed From the Baseline of Dilated ResNet-34.} \label{detection_ablation}
              \centering
              \begin{tabular}{|lcc|}
                  \hline
                  Method    & Proposal Method   & Accuracy (\%)) \\
                  \hline
                  Dilated Resnet-34             & NA &  82.02 \\
                  Degenerate PartNet       & SS &  82.84  \\
                  PartNet             & SS &  \textbf{83.53}  \\
                  % Degenerate PartNet       & DPP & 84.17  \\
                  % PartNet             & DPP & \textbf{84.43}  \\
                  \hline
              \end{tabular}
\end{table}

For the detection stream of PartNet, we need to specify the number $P$ of output neurons of the second FC layer, which is also the specified number of part detectors. To investigate how different values of $P$ influence classification performance, we conduct experiments on the CUB-200-2011 dataset by setting $P = 1, 3, 5$, and $10$. Results in Table \ref{Table:part_detectors} show that classification accuracy slightly improves as more part detectors are used, but at the price of increased computation cost. In our experiments, we set $P = 3$ for a balance between accuracy and efficiency.

            \begin{table}[htbp]
              \caption{Classification Performance on the CUB-200-2011 Dataset \cite{cub2002011} When Using Different Numbers of Part Detectors (i.e., $P$ Values in Eq. (\ref{EqnDetSoftMax1})) in the Detection Stream of PartNet. The PartNet Is Constructed From Dilated ResNet-34.} \label{Table:part_detectors}
              \centering
              \begin{tabular}{|c|cccc|}
                  \hline
                  No. of Part Detectors & 1 & 3 & 5 & 10 \\
                  \hline
                    Accuracy (\%)  & 83.51 & 83.53 & 83.62 & \textbf{83.63}  \\
                  \hline
              \end{tabular}
            \end{table}

	\subsection{Ablation Studies on DPP}
\label{Exp:dpp}

We investigate our proposed DPP method by conducting experiments on the CUB-200-2011 dataset using a PartNet constructed from Dilated ResNet-34.

The number of proposals generated by DPP for each image may influence the performance of PartNet. To investigate, we first generate a number $K = 28$ of boxes for each spatial cell of feature maps (cf. Table \ref{anchor size} in Section \ref{dpp} for how sizes and aspect ratios of the boxes are specified); we then rank the $28$ boxes associated with each cell according to their sizes/areas, and uniformly sample $3$, $7$, and $14$ ones out of them respectively. This creates scenarios of generating $K = 3, 7, 14, 28$ boxes per spatial cell for our proposed DPP. Results in Table \ref{Table:part_proposals} show that classification accuracies slightly improve as more proposals are used. In our experiments, we by default set $K = 28$ for each spatial cell of feature maps.

\begin{table}[htbp]
              \caption{Effect of Different Numbers of Proposals When Using Our Proposed DPP for fine-grained categorization. Experiments Are Conducted on the CUB-200-2011 Dataset \cite{cub2002011} Using a PartNet Constructed From Dilated ResNet-34.} \label{Table:part_proposals}
              \centering
              \begin{tabular}{|c|cccc|}
                  \hline
                  No. of Proposals per Cell & $K = 3$ & $K = 7$ & $K = 14$ & $K = 28$ \\
                  \hline
                    Accuracy (\%)  & 84.31  & 84.41 & 84.36 & \textbf{84.43}  \\
                  \hline
              \end{tabular}
\end{table}

 We also compare with other region proposal methods used in recent fine-grained categorization works \cite{partbased,theapplication}, including SS \cite{selectivesearch} and an improved version of SS termed Filtered SS. Filtered SS removes noisy proposals that are irrelevant to the objects of interest in an image (e.g., those on the background) by an object-level attention model \cite{theapplication}, and it thus enjoys an unfair advantage over both SS and our proposed DPP. For both SS and Filtered SS, we use the same number of region proposals as our DPP does: when these methods produce more proposals, we rank them in terms of areas of proposed regions, and then uniformly sample the same number of region proposals; in some rare case that these methods produce fewer proposals, we also duplicate some ones. Results in Table \ref{proposals} show that our DPP method outperforms both SS and Filtered SS, confirming that candidates of discriminative local parts can be sampled directly at salient positions of feature maps, with no need to be bridged via object-level proposals.

\begin{table}[htbp]
              \caption{Classification Accuracies (\%) of PartNet on the CUB-200-2011 Dataset \cite{cub2002011} When Using Different Methods to Generate Region Proposals.} \label{proposals}
              \centering
              \begin{tabular}{|c|ccc|}
                  \hline
                  Method & SS \cite{selectivesearch} \ & Filtered SS \cite{theapplication} & DPP \\
                  \hline
                  Accuracy (\%)  &83.53  &84.00  & \textbf{84.43}  \\
                  \hline
              \end{tabular}
\end{table}

\begin{table}[htbp]
        \caption{Effect of Resolution of Feature Maps to fine-grained categorization Tasks. The Dilated ResNet-34 Produces Doubled Feature Map Resolution over That of ResNet-34. Experiments Are Conducted on the CUB-200-2011 Dataset \cite{cub2002011}. } \label{Table:stride_size}
        \centering
        \begin{tabular}{|c|c|c|}
        \hline
        ImageNet         & Method                           & Acc. (\%) \\
        Pre-training     &                                   &               \\
        \hline
  		   No          &  ResNet-34                         &  54.44        \\
  		   No          &  Dilated ResNet-34                 &  \textbf{60.95}        \\
  		   \hline
  		   Yes          & ResNet-34                         & 81.78         \\
           Yes          & Dilated ResNet-34                 & \textbf{82.02} \\
           \hline
           Yes         & PartNet constructed from ResNet-34                  & 82.98 \\
           Yes         & PartNet constructed from Dilated ResNet-34          & \textbf{84.36} \\
        \hline
        \end{tabular}
\end{table}

        \begin{table}[htbp]
	\caption{Results of PartNet on the CUB-200-2011 Dataset \cite{cub2002011} With or Without Using Weight Orthogonalization for the Second FC Layer of the Classification Stream.} \label{Table:weight_orthogonal}
	\centering
	\begin{tabular}{|c|c|c|}
		\hline
		Weight Orthogonalization & Method & Accuracy (\%) \\
		\hline
		No          & PartNet                    & 84.43         \\
		Yes         & PartNet                     & \textbf{84.73} \\
		\hline
	\end{tabular}
\end{table}

\subsection{Ablation Studies on Variants of PartNet} \label{Exp_variants}

\begin{table*}[!htb]
	\caption{Classification Accuracies ($\%$) of Individual Models and Various Model Combinations of PartNet With Its Associated Image- and Part-level Models on the CUB-200-2011 \cite{cub2002011}, Oxford Flower 102 \cite{flowers} and Oxford-IIIT Pet \cite{pets} Datasets. % and Oxford Flower 102 \cite{flowers} Datasets.
		%, Stanford Dogs \cite{stanforddog}, Cars-196 \cite{stanfordcar},  Oxford Flowers 102 \cite{flowers} and FGVC-Aircraft \cite{aircraft} datasets.
		The VGGNet Is Used to Extract Features in All the Experiments in This Table and the Symbol "+" Means Averaging the Classification Probabilities of Corresponding Models.} \label{Table:Different_Models}
	\centering
	\begin{tabular}{c|c|c|c}
		\hline
		\hline
		Method              &    CUB-200-2011 \cite{cub2002011}         & Oxford Flower 102 \cite{flowers}     & Oxford-IIIT Pet \cite{pets} \\%       & Oxford Flowers 102      & FGVC-Aricraft \\
		\hline
		Image-level                & 82.19                & 95.12              & 92.07  \\%          & 95.12              & 84.43 \\
		PartNet                    & 85.11                & 95.95              & 92.56  \\ %\         & 95.95              & 84.37 \\
		Part 1                     & 78.51                & 92.79              & 76.97  \\  %\        & 92.79              & \\
		Part 2                     & 75.68                & 93.30              & 83.78  \\ %\         & 93.30              & \\
		Part 3                     & 77.55                & 91.12              & 81.71  \\ %\         & 91.12              & \\
		Part 1 + 2 + 3             & 83.64                & 95.82              & 88.39  \\ %\         & 95.82              & \\
		Part 1 + 2 + 3 + Image-level  & 86.11             & 96.47              & 93.51  \\ %\         & 96.47              & \\
		Image-level + PartNet      & 83.98                & 95.62              & 92.80  \\ %\         & 95.62              & \\
		Part 1 + 2 + 3 + PartNet   & 86.19                & 96.43              & 92.53  \\ %\          & 96.43              & \\
		\hline
		Our PartNet-Full          & \multirow{2}{*}{\textbf{86.90}}       &   \multirow{2}{*}{\textbf{96.70}}    &   \multirow{2}{*}{\textbf{95.37}} \\ %  & \multirow{2}{*}{\textbf{96.70}}  & \multirow{2}{*}{\textbf{xxx}}  \\
		(Part 1 + 2 + 3 + PartNet + Image-level) &  &  & \\
		\hline
		\hline
	\end{tabular}
\end{table*}

Our first PartNet variant is analysed with ResNet-34 and Dilated ResNet-34 models. The Dilated ResNet-34 model uses 2-dilated convolution \cite{dilated} to double the resolution of feature maps without affecting the size of the receptive field. To investigate the effect of feature map resolution itself for fine-grained categorization, we first use the baseline model of ResNet-34, which is either pre-trained on the ImageNet or trained from scratch on the CUB-200-2011 dataset. Since ResNet-34 produces feature maps whose resolution is only half of that of feature maps produced by Dilated ResNet-34, our DPP method cannot generate $K = 28$ per-cell proposals for ResNet-34. We thus set $K = 14$ in this comparative experiment. Results in Table \ref{Table:stride_size} show that for both of the considered training settings (i.e., with or without ImageNet pre-training), higher resolution of feature maps contributes to better classification accuracies, showing its usefulness in fine-grained categorization by preserving finer details of appearance features. When applying dilated convolution to our proposed PartNet (constructed from ResNet-34), performance gets a clear boost as well.

Our second PartNet variant enforces weight matrix of the second FC layer in the classification stream to be orthogonal, by using the SVB technique proposed in \cite{svb}. To investigate its effectiveness, we again conduct experiments on the CUB-200-2011 dataset using PartNet constructed from Dilated ResNet-34. Results in Table \ref{Table:weight_orthogonal} show that this variant achieves improved classification performance. Note that results of PartNet and PartNet-Full reported in Sections \ref{Exp_Different_Models} and \ref{Exp_Comparison} are based on this variant.

\begin{table*}[!htb]
	\caption{Classification Results of Different Methods on the CUB-200-2011 \cite{cub2002011} Dataset.}\label{Table:cub comparison}
	\centering
	\begin{tabular}{lcccc}
		\hline
		\hline
		Method                                     & Training annotation & Test annotation  & CNN Features  & Accuracy (\%)  \\  %& Architecture
		\hline
		
		VGG-BGLm \cite{vgg-bglm}                    & BBox             & BBox               & VGGNet      & 80.40   \\
		PG Alignment \cite{pd}                      & BBox             & --                 & VGGNet      & 82.00   \\ % & VGGNet
		Coase-to-Fine \cite{coarse-to-fine}         & BBox             & --                 & VGGNet      & 82.50   \\
		PG Alignment \cite{pd}                      & BBox             & BBox               & VGGNet      & 82.80   \\
		Coase-to-Fine \cite{coarse-to-fine}         & BBox             & BBox               & VGGNet      & 82.90   \\
		PBC \cite{pbc}                             & BBox             & --                 & GoogleNet   & 83.30   \\
		FCAN \cite{fcan}                            & BBox             & BBox               & ResNet-50   & 84.70   \\
		\hline
		
		Part-based RCNN \cite{partbased}           & BBox + Parts     & --                 & AlexNet      & 73.90  \\ %& AlexNet
		PBC \cite{pbc}                              & BBox + Parts     & BBox               & GoogleNet    & 83.70  \\
		DPS-CNN \cite{realtime}                    & Parts            & --                 & GoogleNet    & 85.12  \\
		SPDA \cite{spdacnn}                        & BBox + parts     & BBox               & VGGNet       & 85.14  \\ %& VGGNet
		Zhang et al. \cite{finegrainedpose}        & Parts            & --                 & VGGNet       & 85.92  \\
		HSnet \cite{hsnet}                         & Parts            & --                 & GoogleNet    & \textbf{87.50}  \\
		\hline
		\hline
		
		& & & & \\
		
		\hline
		\hline
		Two-level Attention \cite{theapplication}  & --               & --                 & AlexNet     & 69.70   \\
		VGG-BGLm \cite{vgg-bglm}                   & --               & --                 & VGGNet      & 75.90   \\
		DVAN \cite{dvan}                           & --               & --                 & VGGNet      & 79.00   \\
		Zhang \emph{et al.} \cite{weaklysupervised}       & --               & --                 & VGGNet      & 79.34  \\
		NAC \cite{nac}                             & --               & --                 & VGGNet      & 81.01   \\
		STN \cite{stn}                             & --               & --                 & GoogleNet   & 84.10   \\
		Bilinear-CNN \cite{bilinear}             & --                & --                  & VGGNet      & 84.10   \\
		FCAN \cite{fcan}                           & --               & --                 & ResNet-50   & 84.30  \\
		PDFS \cite{pdfs}                           & --               & --                 & VGGNet      & 84.54  \\
		PNA \cite{picking_journal} & --               & --                 & VGGNet      & 84.70  \\
		RA-CNN \cite{lookcloser}                   & --               & --                 & VGGNet      & 85.30   \\
		MA-CNN (2 parts + object) \cite{macnn}     & --               & --                 & VGGNet      & 85.40   \\
		OPAM \cite{opaddl}                       & --               & --                   & VGGNet      & 85.83  \\
		DT-RAM \cite{dt-ram}                       & --               & --                 & ResNet-50   & 86.00   \\
		%\textcolor {red} {Sun \emph{et al.} \cite{mamc}}                     & --               & --                 & ResNet-50   & 86.20  \\
		MA-CNN (4 parts + object) \cite{macnn}     & --               & --                 & VGGNet      & 86.50  \\
		%\textcolor {red} {DFL-CNN \cite{dfl}}                         & --               & --                 & VGGNet      & 86.70  \\
		%\textcolor {red} {NTS-Net \cite{nts}}                         & --               & --                 & ResNet-50   & \textbf{87.50} \\
		
		\hline
		
		Our PartNet-Full                               & --               & --                 & VGGNet      & \textbf{86.90} \\
		Our PartNet-Full                               & --               & --                 & ResNet-34      & \textbf{87.30} \\
		%\textcolor {red} {Our PartNet-Full}                               & --               & --                 & ResNet-50      & \textcolor {red} {\textbf{xxxx}} \\
		\hline
		\hline
		
	\end{tabular}
\end{table*}

\begin{figure}[!htb]
	\begin{center}
		\centering
		\includegraphics[width=0.95\linewidth]{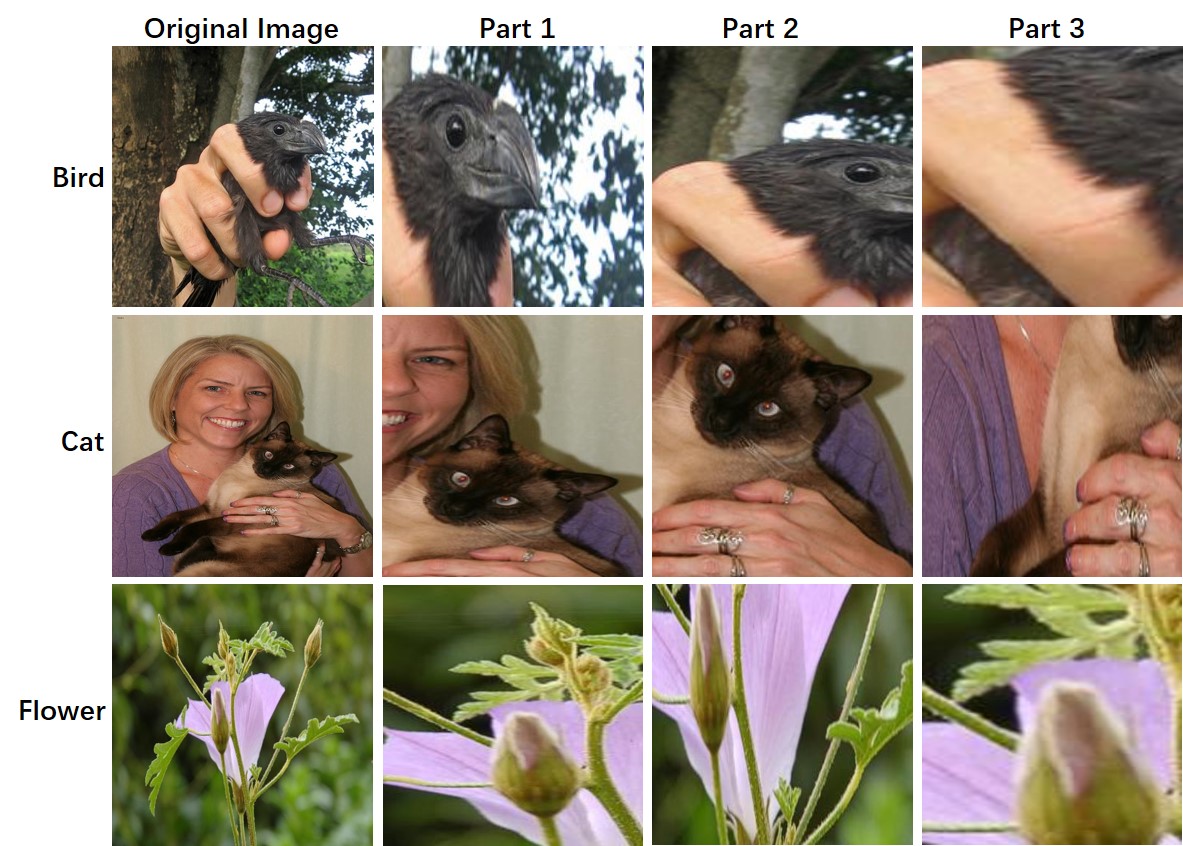}
	\end{center}
	\caption{Some failure results of part detection. Images of ``Bird'', ``Cat'', and ``Flower'' are from CUB-200-2011 \cite{cub2002011}, Oxford-IIIT Pet \cite{pets}, and Oxford Flower 102 \cite{flowers} datasets, respectively.}
	\label{fig:detection_error}
\end{figure}

\subsection{Ensemble of PartNet with Its Associated Image- and Part-level Models} \label{Exp_Different_Models}

We introduce in Section \ref{implementation details} that PartNet is constructed from an image-level base model, and multiple part-level models can also be obtained by fine-tuning the image-level model with the region proposals respectively detected by the learned part detectors of PartNet. PartNet contributes to fine-grained categorization by aggregating local discriminative evidence provided by part detectors. Complementary to PartNet, image- and part-level models may respectively provide their own discrimination by emphasizing either the holistic image or each of the individual parts. It is arguably beneficial to use an ensemble of these models to further boost classification performance. Empirical success of similar model ensemble is also presented in {\cite{opaddl, macnn, lookcloser}.

For model ensemble in this section, we use the VGGNet based PartNet with the two variants introduced in Section \ref{SecPartNetVariants}. Table \ref{Table:Different_Models} shows results of individual models and various model combinations of the ensemble. We can observe that:

\begin{itemize}
	\item Classification accuracies of the individual part-level models are relatively low, which may be attributed to the relatively less information contained in individual parts, and also the occasional failure of part detection. Figure \ref{fig:detection_error} shows two main reasons (i.e., complex background and heavy occlusion) causing the failure of part detection. However, averaging the predictions of three part-level models boosts the accuracy by a large margin, proving that the detected three part-level regions are complementary with each other.
	\item Averaging the classification probabilities of image- and part-level models achieves large performance improvements (e.g., 3.92\%, 1.35\%, and 1.44\% on the datasets of CUB-200-2011, Oxford Flower 102, and Oxford-IIIT Pet respectively) over the results of using the image-level models alone, justifying the complementarity of holistic image and individual parts.
	\item Combining PartNet, image-level, and part-level models further boosts classification performance on the three datasets, certifying the effectiveness of our proposed method.
\end{itemize}
\color{black}

         \begin{figure*}[!htb]
	\begin{center}
		
		%\begin{minipage}[b]{0.4\linewidth}
		\subfigure[Birds]{
			\centering
			\includegraphics[width=0.9\linewidth]{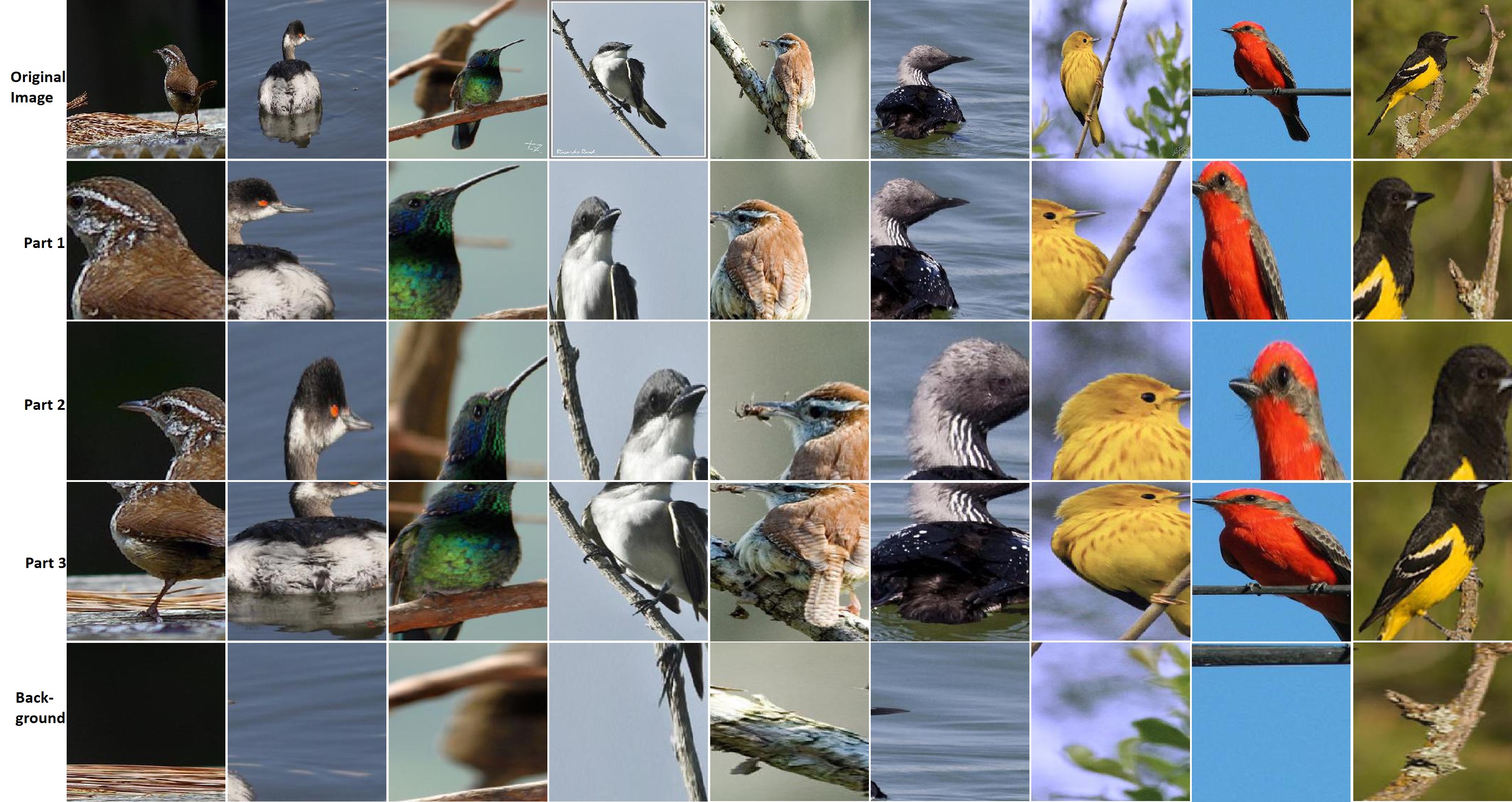}
		}
		%\end{minipage}
		%\hfill
		%\begin{minipage}[b]{0.4\linewidth}	
		%        		\subfigure[Cars]{
		%         	%\begin{minipage}[c]{0.8\textwidth}
		%         	    \centering
		%        	    \includegraphics[width=0.459\linewidth]{fine_grained_image//flowers.jpg}
		%         	%\end{minipage}
		%        		}
		%\end{minipage}
		\subfigure[Flowers]{
			\centering
			\includegraphics[width=0.48\linewidth]{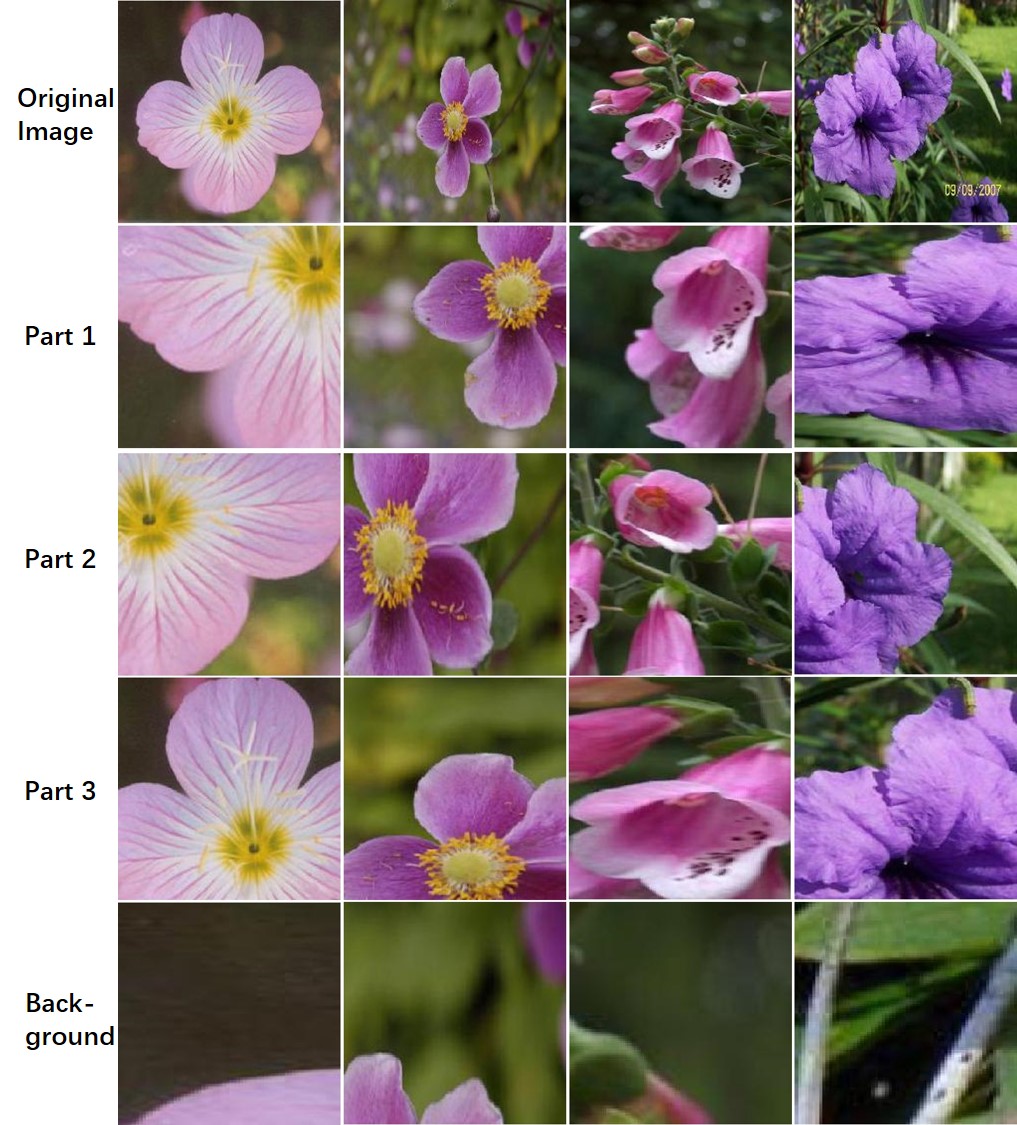}
		}
		\hfill
		\subfigure[Pets]{
			\centering
			\includegraphics[width=0.48\linewidth]{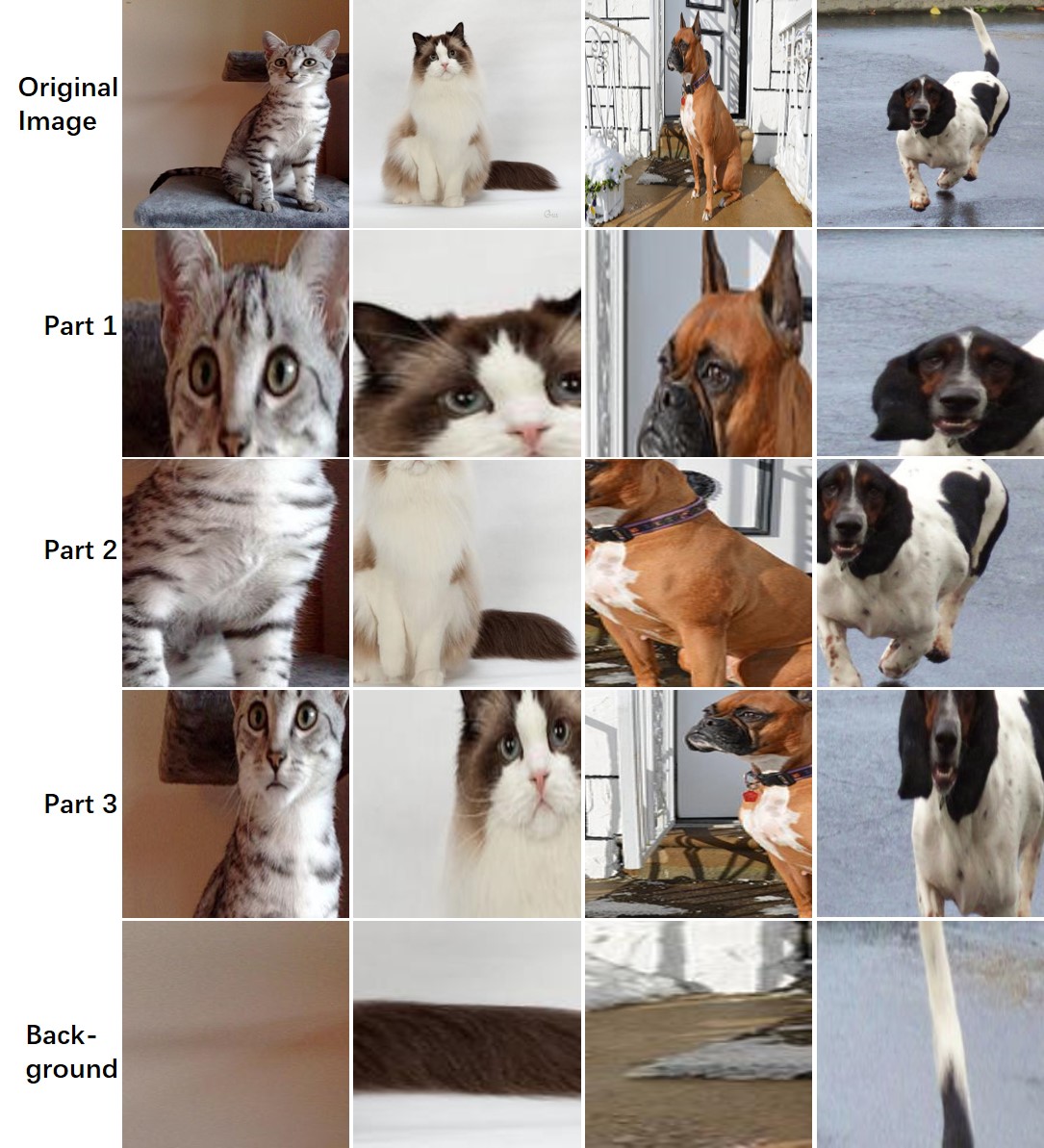}
		}
		%        		\subfigure[Dogs]{
		%         	    \centering
		%        	    \includegraphics[width=0.3\linewidth]{fine_grained_image//dogs_three.jpg}
		%        		}
		%        		\subfigure[Aircraft]{
		%         	    \centering
		%        	    \includegraphics[width=0.3\linewidth]{fine_grained_image//dogs_three.jpg}
		%\includegraphics[width=0.8\linewidth]{fine_grained_image//final-bird.jpg}
	\end{center}
	\caption{Visualization of the detected parts on datasets of CUB-200-2011 \cite{cub2002011}, Oxford Flower 102 \cite{flowers} and Oxford-IIIT Pet \cite{pets}. The first row denotes the original images, and the second, third, and fourth rows denote the parts detected by the three part detectors respectively. The last row denotes the background or less discriminative proposals detected by the background detector. Results are obtained by the PartNet constructed from the VGGNet with the two variants introduced in Section \ref{SecPartNetVariants}. The images in (a) Birds, (b) Flowers and (c) Pets are from the test data of CUB-200-2011 \cite{cub2002011}, Oxford Flower 102 \cite{flowers} and Oxford-IIIT Pet \cite{pets} datasets respectively.}
	%The images in (a) Birds, (b) Cars, (c) Flowers, (d) Dogs, (e) Aircraft,  are from the test set of CUB-200-2011 \cite{cub2002011}, Cars-196 \cite{stanfordcar}, Oxford Flowers 102 \cite{flowers}, Stanford Dogs \cite{stanforddog} and FGVC-Aircraft \cite{aircraft} datasets respectively. (best view in color). \textcolor {red} {TO BE MODIFIED}}
	\label{fig:part detector}
	
\end{figure*}

             \begin{table}[!htb]
              \caption{Classification Results of Different Methods on the Oxford Flower 102 \cite{flowers} Dataset.}\label{Table:Flowers102 Comparison}
              \centering
              \begin{tabular}{|lcc|}
                  \hline
                   Method                          & CNN Features  & Accuracy (\%)  \\  %& Architecture
                  \hline
                  MPP \cite{mpp}                    & AlexNet       & 91.28   \\
                  Magnet \cite{magnet}              & GoogleNet     & 91.40   \\
                  BoSP \cite{surrogate-part-features} & VGGNet      & 94.02   \\
                  NAC \cite{nac}                    & VGGNet        & 95.34   \\
                  PBC \cite{pbc}                    & GoogleNet     & 96.10   \\
                  OPAM \cite{opaddl}                & VGGNet        & \textbf{97.10} \\
                  \hline
                  Our PartNet-Full                  & VGGNet      & \textbf{96.70} \\			
                  \hline
              \end{tabular}
            \end{table}

        \begin{table}[!htb]
        	\caption{Classification Results of Different Methods on the Oxford-IIIT Pet \cite{pets} Dataset.}\label{Table:Pet_Comparison}
        	\centering
        	\begin{tabular}{|lcc|}
        		\hline
        		Method                          & CNN Features  & Accuracy (\%)  \\  %& Architecture
        		\hline
        		NAC \cite{nac}                    & VGGNet        & 91.60   \\
        		Two-level Attention \cite{theapplication} & VGGNet& 92.51   \\
        		OPAM \cite{opaddl}                & VGGNet        & \textbf{93.81} \\
        		\hline
        		Our PartNet-Full                  & VGGNet      & \textbf{95.37} \\			
        		\hline
        	\end{tabular}
        \end{table}

\subsection{Comparison with State-of-the-art Methods} \label{Exp_Comparison}

We compare our PartNet with the state-of-the-art methods on the benchmark datasets of CUB-200-2011 \cite{cub2002011}, Oxford Flower 102 \cite{flowers} and Oxford-IIIT Pet \cite{pets}.
%5 widely-used fine-grained datasets.
Table \ref{Table:cub comparison} presents comparison results on the CUB-200-2011 dataset. The types of annotation used in the training and test stages of each method are also listed in the table, where ``CNN Features'' indicates which (base) network is used to extract features in each method.

When constructing the PartNet using the VGGNet (with the two variants introduced in Section \ref{SecPartNetVariants}), our PartNet-Full (i.e., the ensemble model described in Section \ref{Exp_Different_Models}) obtains the new state-of-the-art result on the CUB-200-2011 dataset when neither object nor part annotations are used. Furthermore, our method outperforms most of the existing ones that need part or object annotations, such as \cite{spdacnn, finegrainedpose, realtime, partbased}. When constructing the PartNet using the base network of Dilated ResNet-34, our PartNet-Full obtains an even better result on the CUB-200-2011 dataset.

Note that the state-of-the-art method HSnet \cite{hsnet} uses the ground-truth part annotations in the training stage, making it less relevant to compare directly with our proposed method. Our PartNet-Full combines multi-level models for final prediction by simply averaging the classification probabilities of these models. In contrast, the MA-CNN \cite{macnn} trains a classifier based on the concatenated features of multi-level models, and OPAM \cite{opaddl} learns a weight for each model with the computationally expensive k-fold cross-validation method, yet their results are still worse than ours.

We present our result on the Oxford Flower 102 dataset in Table \ref{Table:Flowers102 Comparison}. Our PartNet-Full obtains the result that is comparable with state-of-the-art method \cite{opaddl}.

We also present our result on the Oxford-IIIT Pet dataset in Table \ref{Table:Pet_Comparison}. Our PartNet-Full obtains the new state-of-the-art result, justifying the efficacy of our PartNet.
%The PartNet in this section is built on the VGGNet with the two variants introduced in Section \ref{SecPartNetVariants} and the PartNet-Full is obtained in the way described in Section \ref{Sec_Final_Prediction}. Our PartNet-Full obtains the new state-of-the-art result on the CUB-200-2011 dataset when neither object nor part annotations are used. Furthermore, our approch outperforms most of the methods that need part or object annotations, such as \cite{spdacnn, finegrainedpose, realtime}, \cite{partbased}.

%Note that the state-of-the-art method HSnet \cite{hsnet} not only uses the ground-truth part annotations in the training stage, but also is constructed on GoogleNet \cite{googlenet}, which is a more powerful network structure than the widely-used VGGNet. So the HSnet enjoys unfair advantages than our approach. We combine multi-level models for final prediction by simply averaging the classification probabilities of those models. In contrast, the MA-CNN \cite{macnn} trains an extra classifier based on the concatenated features of multi-level models and OPAM \cite{opaddl} learns a weight for each model with the computationally expensive $k$-fold cross validation method.

\subsection{Part Detection Visualization} \label{subsec:part_visalization}

      %We visualize the detected discriminative parts on the PartNet constructed from the VGGNet with the two variants introduced in Section \ref{SecPartNetVariants}. The detected part-level proposals of the CUB-200-2011 \cite{cub2002011} and Oxford Flower 102 \cite{flowers} datasets are presented in Figure \ref{fig:part detector} and the illustrated images are from the test data of corresponding datasets. The detected part regions for each dataset are roughly consistent with some other methods \cite{opaddl}, \cite{partbased} (e.g., the Part 2 and Part 3 are detected as the head and body of birds in the CUB-200-2011 dataset). What's more, our PartNet also detects one other discriminative part (e.g., the detected Part 1 for the birds). The detector for background regions gathers the background proposals that are unrelated with the objects. In this way, the influences of background proposals for image category prediction can be decreased by removing the background detector before combining the two streams (cf. Section \ref{Sec_aggregate_two_branches}).

      In Figure \ref{fig:part detector}, we visualize the detected discriminative parts by the VGGNet based PartNet (with the two variants introduced in Section \ref{SecPartNetVariants}), where images are from test data of the CUB-200-2011 \cite{cub2002011}, Oxford Flower 102 \cite{flowers} and Oxford-IIIT Pet datasets. We observe in Figure \ref{fig:part detector} that our detected local parts have physical meanings:

\begin{itemize}
	\item For the bird dataset, the first two parts (Part 1 and Part 2) are on local regions of bird head, with the second one being a slightly zoomed-in version of the first one, and the third part (Part 3) is on local regions of bird body (back and/or abdomen).
	\item For the flower dataset, the three parts are roughly on local regions of a flower or some of its petals, regardless of how many flowers are contained in each of the images.
	\item For the pet dataset, the first part (Part 1) and the third part (Part 3) are on local regions of pet head, with the first one being a slightly zoomed-in version of the third one, and the second part (Part 2) is on local regions of pet body.
\end{itemize}
\color {black}
 %for the bird dataset, the first two parts (Part 1 and Part 2) are on local regions of bird head, with the second one being a slightly zoomed-in version of the first one, and the third part (Part 3) is on local regions of bird body (back and/or abdomen); for the flower dataset, the three parts are roughly on local regions of a flower or some of its petals, regardless of how many flowers are contained in each of the images; the part detection of the pet dataset is similar to that of the bird dataset: the first part (Part 1) and third part (Part 3) are on local regions of pet head, with the first one being a slightly zoomed-in version of the third one, and the second part (Part 2) is on local regions of pet body.
These detected local parts arguably provide semantically discriminative information for fine-grained categorization. Figure \ref{fig:part detector} also shows that the background detector gathers region proposals that are on the image background and are thus less relevant to the task of interest. The influence of background proposals for image category prediction can thus be decreased by removing the background detector before combining the two streams (cf. Section \ref{Sec_aggregate_two_branches}).

\section{Conclusions}
     In this paper, we propose a novel Weakly Supervised Part Detection Network (PartNet) for part-aware fine-grained object categorization. Our PartNet contains two streams: the classification stream classifies part-level region proposals over subordinate categories; the detection stream selects discriminative proposals for the use of fine-grained object categorization. The image-level classification is obtained by the combination of region-level probabilities of the two streams, and meanwhile diverse part detectors can be learned in an end-to-end fashion under the image-level supervision. To prepare part-level region proposals for the PartNet, we design a simple Discretized Part Proposals method that utilizes the localization information in the feature maps directly. Experiments on the benchmark datasets of CUB-200-2011 \cite{cub2002011}, Oxford Flower 102 \cite{flowers} and Oxford-IIIT Pet \cite{pets} demonstrate the efficacy of our proposed PartNet on fine-grained categorization and salient part detection. Especially our approach obtains the new state-of-the-art result on the CUB-200-2011 and Oxford-IIIT Pet datasets when ground-truth part annotations are not available. We believe that such methods, that only need image categorization level supervision are important for new fine-grained categorization tasks.

% or



\ifCLASSOPTIONcaptionsoff
  \newpage
\fi

\begin{IEEEbiography}[{\includegraphics[width=1in,height=1.25in,clip,keepaspectratio]{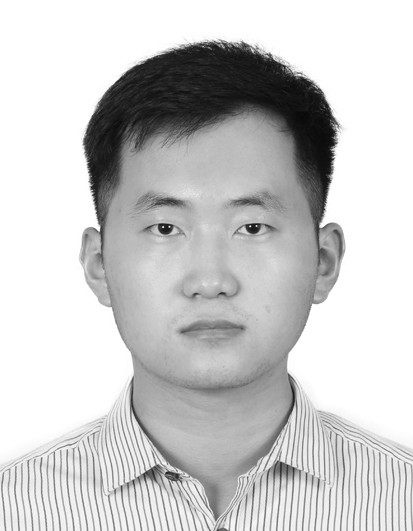}}]{Yabin Zhang}
received the B.E. degree in School of Electronic and Information Engineering from South China University of Technology, Guangzhou, China, in 2017, where he is currently pursuing the master's degree. His current research interests include computer vision and deep learning, especially the deep transfer learning.
\end{IEEEbiography}

\begin{IEEEbiography}[{\includegraphics[width=1in,height=1.25in,clip,keepaspectratio]{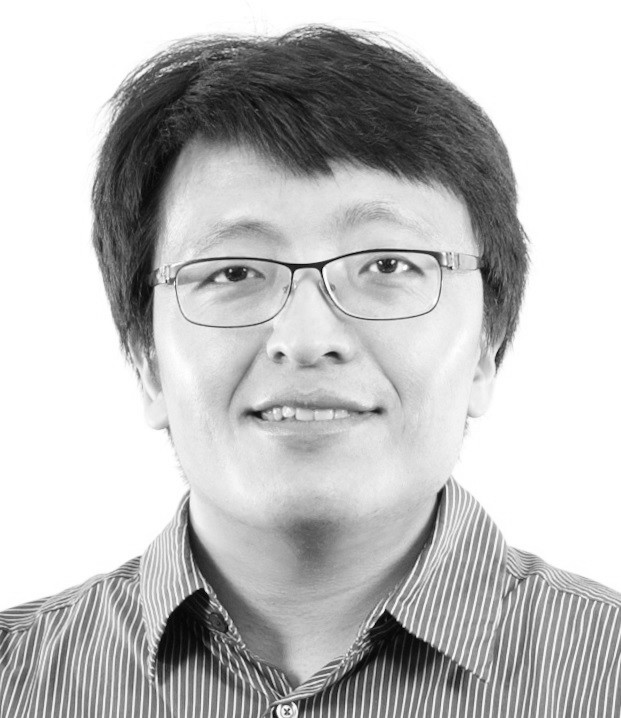}}]{Kui Jia} received the B.E. degree from Northwestern Polytechnic University, Xi’an, China, in 2001, the M.E. degree from the National University of Singapore, Singapore, in 2004, and the Ph.D. degree in computer science from the Queen Mary University of London, London, U.K., in 2007.

He was with the Shenzhen Institute of Advanced Technology of the Chinese Academy of Sciences, Shenzhen, China, Chinese University of Hong Kong, Hong Kong, the Institute of Advanced Studies, University of Illinois at Urbana-Champaign, Champaign, IL, USA, and the University of Macau, Macau, China. He is currently a Professor with the School of Electronic and Information Engineering, South China University of Technology, Guangzhou, China. His recent research focuses on theoretical deep learning and its applications in vision and robotic problems, including deep learning of 3D data and deep transfer learning. 
\end{IEEEbiography}

\begin{IEEEbiography}[{\includegraphics[width=1in,height=1.25in,clip,keepaspectratio]{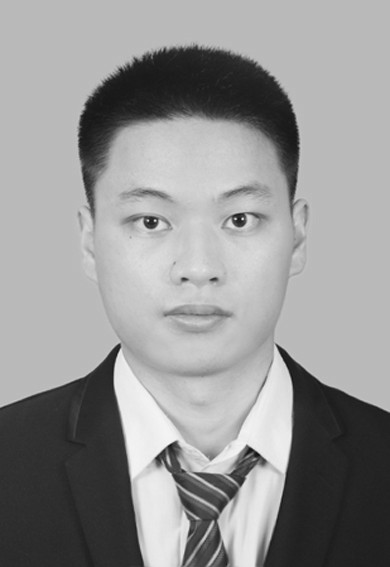}}]{Zhixin Wang}
received the B.E. degree in School of Electronic and Information Engineering from South China University of Technology, Guangzhou, China, in 2017, where he is currently pursuing the master's degree. His recent research focuses on computer vision and deep learning, especially the object detection.
\end{IEEEbiography}


%\vfill

{\small
\bibliographystyle{IEEETran}
\bibliography{egbib}
}

\end{document}